\documentclass[a4paper,fleqn]{cas-sc}

\usepackage{amsmath}
\usepackage{amssymb}
\usepackage{booktabs}
\usepackage{multirow}
\usepackage{placeins}
\usepackage{float}

\usepackage{capt-of}
\usepackage{algorithm}
\usepackage{algpseudocode}
\usepackage[numbers]{natbib}
\hypersetup{
    colorlinks=true,
    citecolor=blue,
    linkcolor=blue,
    urlcolor=blue,
    pdftitle={FAST-DeepONet: Factor-Augmented Branch Representations for High-Dimensional PDE Inputs in the Small-Sample Regime},
    pdfauthor={Jiyong Kwon; BongSeok Kim; Guang Lin},
    pdfkeywords={DeepONet, operator learning, dimension reduction}
}

\graphicspath{{figures/}}

\newcommand{\R}{\mathbb{R}}
\newcommand{\calD}{\mathcal{D}}
\newcommand{\calG}{\mathcal{G}}
\newcommand{\calX}{\mathcal{X}}

\newcommand{\hatf}{\widehat f}

\begin{document}
\let\WriteBookmarks\relax

\shorttitle{FAST-DeepONet for High-Dimensional PDEs}
\shortauthors{J. Kwon et~al.}

\title[mode=title]{FAST-DeepONet: Factor-Augmented Branch Representations for High-Dimensional PDE Inputs in the Small-Sample Regime}

\author[1]{Jiyong Kwon}[orcid=0009-0003-4426-3419]
\ead{kwon165@purdue.edu}
\credit{Conceptualization, Methodology, Software, Formal analysis, Investigation, Writing -- original draft}

\author[1]{BongSeok Kim}[orcid=0009-0000-9436-438X]
\ead{kim4853@purdue.edu}
\credit{Data curation, Methodology, Investigation, Writing -- review and editing}

\author[1,2]{Guang Lin}[orcid=0000-0002-0976-1987]
\cormark[1]
\ead{guanglin@purdue.edu}
\credit{Supervision, Conceptualization, Writing -- review and editing}

\cortext[1]{Corresponding author.}

\affiliation[1]{
    organization={School of Mechanical Engineering, Purdue University},
    addressline={585 Purdue Mall},
    city={West Lafayette},
    postcode={47907},
    state={IN},
    country={USA}
}

\affiliation[2]{
    organization={Department of Mathematics, Purdue University},
    addressline={150 N. University Street},
    city={West Lafayette},
    postcode={47907},
    state={IN},
    country={USA}
}

\begin{abstract}
Deep operator networks can become statistically unstable when partial differential equation inputs are observed at thousands of strongly correlated sensors but only a small number of operator samples is available. We introduce FAST-DeepONet, a branch representation combining a fixed spectral path with a regularized projection of the orthogonal residual, in which the directional penalty acts on the effective residual map after each of its rows is normalized. On Navier--Stokes flow a plain DeepONet degrades from $0.0394$ to $0.1556$ mean relative $L_2$ error as the branch grows from $129$ to $8193$ coordinates, while FAST-DeepONet stays near $0.04$, so the sensor grid can be refined without a statistical penalty. Across independent test sets for Navier--Stokes flow, Darcy flow, and signed terminal wavefield prediction it lowers mean relative $L_2$ error by $4.7\%$ to $37.0\%$ with three to seven times fewer trainable parameters. A spectral-only branch sharing the same basis separates the two paths: the fixed spectral path carries the improvement on Navier--Stokes and Darcy, while terminal wave prediction requires the residual path together with its directional penalty. FAST-DeepONet targets coordinate-query architectures and trains on solution values alone.
\end{abstract}


\maketitle
\hypersetup{pdfauthor={Jiyong Kwon; BongSeok Kim; Guang Lin}}

\section{Introduction}

Parametric partial differential equations (PDEs) arise in climate modeling, subsurface flow, energy systems, and engineering design. Many applications require repeated solutions as coefficients, initial conditions, or physical parameters vary. Operator learning replaces repeated numerical solves by learning maps between input and output functions. DeepONet \citep{lu2021deeponet}, motivated by operator approximation results \citep{chen1995universal}, represents the input function with a branch network and the output coordinate with a trunk network. The branch input is commonly a coefficient or initial field sampled at a fixed sensor layout. A fine discretization can produce thousands of strongly dependent covariates, even when only tens of independent PDE solutions are available for training. A raw branch must then estimate its first layer in a regime where the input dimension can greatly exceed the number of operator samples.

Reduced representations are a natural response, and the existing ones act on different parts of the architecture. POD-DeepONet replaces the trunk with fixed modes computed from the output fields, so it compresses the solution side and leaves the branch input untouched \citep{lu2022comprehensive}. PCA-Net reduces both the input and the output to principal component coefficients and learns a map between them, without a coordinate query head \citep{bhattacharya2021model}. Latent-space operator learning compresses both sides \citep{kontolati2024learning} and NOMAD replaces the linear output decoder \citep{seidman2022nomad}, so both alter the output construction we keep fixed. DE-DeepONet acts on the branch input itself, using a derivative-informed subspace together with derivative supervision \citep{qiu2024derivative}, which requires access to derivatives of the solution operator. The present work acts on the branch input but uses only solution values for supervision, so derivative-informed subspaces are unavailable here. We therefore ask whether a structured input representation estimated from solution data alone can stabilize the DeepONet branch while leaving the trunk and output construction unchanged.

FAST-NN, the factor-augmented sparse-throughput neural network of Fan and Gu \citep{fan2024fastnn}, combines common factor features with a trainable path for information outside the factor subspace in regression with high-dimensional inputs. We adapt this idea to the branch network of DeepONet. FAST-DeepONet represents each input using spectral factors estimated from the training data and a learned projection of the orthogonal residual into a small number of features. In our experiments the residual projection places above-chance weight on the coordinates that carry the localized input structure. The selected coordinates nonetheless vary across fits and cover only part of that structure, so we read the path as a residual representation rather than as sparse variable selection.

The adaptation raises a parameterization issue. A direct penalty on the residual projection would depend on how adjacent trainable layers are scaled, even when they represent the same input map. It could also penalize components that are removed by the orthogonal residual projector. We therefore regularize the effective residual map after normalizing each of its rows.

The main contributions of this work are:
\begin{enumerate}
    \item We introduce a DeepONet branch that combines a fixed spectral path, a numerical rank safeguard, and a trainable projection of the orthogonal residual. In the present experiments this converts 4,096 to 8,193 sensor values into a branch representation of at most 32 features. The first branch layer shrinks with it, so the model trains three to seven times fewer parameters than a plain branch.

    \item We identify the effective residual map as the appropriate object to regularize and apply a directional penalty after exact row normalization. This construction addresses the scaling and factor subspace ambiguities of a direct penalty on the projection parameters without assuming sparse coordinate recovery. An ablation shows that the penalty is what makes the residual path useful.

    \item We evaluate FAST-DeepONet on independent test sets for Navier--Stokes flow, Darcy flow, and signed terminal wavefield prediction, where it reduces mean relative $L_2$ error by $4.7\%$ to $37.0\%$ relative to a plain branch. A sensor and sample study on all three benchmarks shows that the advantage grows with the branch dimension at fixed sample size, and a spectral-only reference identifies which of the two paths carries it.
\end{enumerate}

The main comparison keeps the DeepONet trunk and output construction fixed, so only the branch representation changes. A spectral-only branch that shares the same basis but omits the residual path serves as the input-side reduced-basis reference throughout.

\section{Preliminaries}
\label{sec:prelim}

\subsection{Operator learning with DeepONet}

Operator learning seeks to approximate mappings between function spaces from paired observations of input and output functions \citep{kovachki2023neural}. In a parametric PDE, such a mapping is induced by the solution process. Consider a family of equations written abstractly as
\[
    \mathcal{F}(s;a,\eta)=0,
\]
where $a\in\calX$ denotes a spatially varying input and $\eta\in\R^{p_\eta}$ contains additional physical parameters. Solving the equation defines an operator
\[
    \calG:\calX\times\R^{p_\eta}\rightarrow\mathcal{Y},
    \qquad
    (a,\eta)\mapsto \calG(a,\eta),
\]
where $\mathcal{Y}$ is the output function space. For an evolution problem, $a$ may specify the initial state and the operator may return the terminal solution
\[
    \calG_T(a,\eta)=s(T,\cdot).
\]
For an elliptic problem, $a$ may represent a spatial coefficient and $\calG(a,\eta)$ is the corresponding stationary solution.

DeepONet represents the input function through samples collected at a fixed set of sensor locations \citep{lu2021deeponet}. Suppose that $a:\Omega\rightarrow\R^{d_a}$ is observed at sensors $\{\zeta_k\}_{k=1}^{m}$. The branch input is
\begin{equation}
    x=
    \bigl[
        a(\zeta_1)^\top,\ldots,
        a(\zeta_m)^\top,
        \eta^\top
    \bigr]^\top
    \in\R^p,
    \qquad
    p=md_a+p_\eta.
    \label{eq:branch_vector}
\end{equation}
This formulation includes scalar coefficient fields as well as vector-valued initial conditions.

For an output coordinate $\xi\in\R^d$, DeepONet approximates the solution operator using
\begin{equation}
    \widehat{\calG}(a,\eta)(\xi)
    =
    \sum_{\ell=1}^{q}
    b_\ell(x)t_\ell(\xi)+c,
    \label{eq:deeponet}
\end{equation}
where the branch network $b$ encodes the sampled input and the trunk network $t$ encodes the output coordinate. This separable representation is motivated by universal approximation results for nonlinear operators \citep{chen1995universal,lu2021deeponet}, for which error estimates are available \citep{lanthaler2022error}.

The input sensor locations and their ordering are fixed across samples in the present study. They do not need to coincide with the output coordinates supplied to the trunk network. The output coordinates are normalized before entering the trunk, and all reported errors are evaluated on the native solver grids. Accuracy away from these grids is not considered here.

\subsection{Factor augmented regression and FAST-NN}

Approximate factor models describe a high-dimensional covariate as the sum of common and idiosyncratic components \citep{stock2002forecasting,bai2002determining,fan2013poet}:
\begin{equation}
    x=Bf+u,
    \qquad
    B\in\R^{p\times r_0},
    \quad
    f\in\R^{r_0},
    \quad
    u\in\R^p.
    \label{eq:factor_model}
\end{equation}
Here, $Bf$ captures variation shared across many coordinates, while $u$ contains variation not explained by the common factors. The FAST regression model, in the lineage of factor-adjusted model selection \citep{fan2020farm}, assumes that the response depends on the latent factors and a small subset of the idiosyncratic coordinates:
\begin{equation}
    y=m^*(f,u_J)+\varepsilon,
    \qquad
    u_J=(u_j:j\in J),
    \qquad
    |J|\ll p.
    \label{eq:fast_response}
\end{equation}

Because $f$ is not observed, FAST-NN uses a diversified projection $W_{\mathrm{DP}}\in\R^{p\times r}$ to construct factor features \citep{fan2022learning,fan2024fastnn}:
\begin{equation}
    \widehat f
    =
    \frac{1}{p}W_{\mathrm{DP}}^\top x.
    \label{eq:diversified_projection}
\end{equation}
Substituting \eqref{eq:factor_model} gives
\begin{equation}
    \widehat f
    =
    Hf+\varsigma,
    \qquad
    H=\frac{1}{p}W_{\mathrm{DP}}^\top B,
    \qquad
    \varsigma=\frac{1}{p}W_{\mathrm{DP}}^\top u.
    \label{eq:diversified_decomposition}
\end{equation}
When the projection is sufficiently aligned with the factor loading space and the idiosyncratic components are weakly dependent, $Hf$ preserves the common factor information while the contribution of $\varsigma$ decreases as the ambient dimension grows.

FAST-NN combines these factor features with a trainable throughput path:
\begin{equation}
    m_{\mathrm{FAST}}(x)
    =
    g\left(
        \left[
            \widehat f,\,
            \overline T_M(\Theta^\top x)
        \right]
    \right),
    \label{eq:fastnn_predictor}
\end{equation}
where $g$ is a ReLU network, $\Theta\in\R^{p\times s}$ is a trainable variable selection matrix, and $\overline T_M$ is a truncation operator. A clipped penalty on the entries of $\Theta$ \citep{fan2001variable} encourages the network to use a small number of additional input coordinates. Under the factor, dependence, and sparsity assumptions of Fan and Gu, the resulting error bounds are governed mainly by the intrinsic factor and sparse dimensions rather than by a generic nonparametric rate in $p$ dimensions.

FAST-DeepONet retains the two-path structure but adapts it to operator learning. The fixed path uses spectral features estimated from the training inputs, while the trainable path acts explicitly on the orthogonal residual. The scalar regression network is replaced by a DeepONet branch whose output is combined with a coordinate-dependent trunk network. The FAST-NN theory motivates this construction, but its scalar regression guarantees do not directly extend to the operator setting considered here.

We keep the name FAST as a record of this lineage. The throughput path and its clipped penalty are retained, but act on the orthogonal residual and are regularized through the effective map of Section~\ref{sec:method}. That map concentrates on a small fraction of the branch coordinates, though not on the same ones across fits, so we describe it as a distributed residual representation; Appendix~\ref{app:map_diagnostics} quantifies both.

\section{FAST-DeepONet}
\label{sec:method}

FAST-DeepONet replaces the raw high-dimensional branch input with fixed spectral coordinates and a learned low-dimensional encoding of the orthogonal residual. Figure~\ref{fig:architecture} summarizes the construction.

\par\medskip
\noindent
\begin{minipage}{\linewidth}
    \centering
    \includegraphics[width=\linewidth]
        {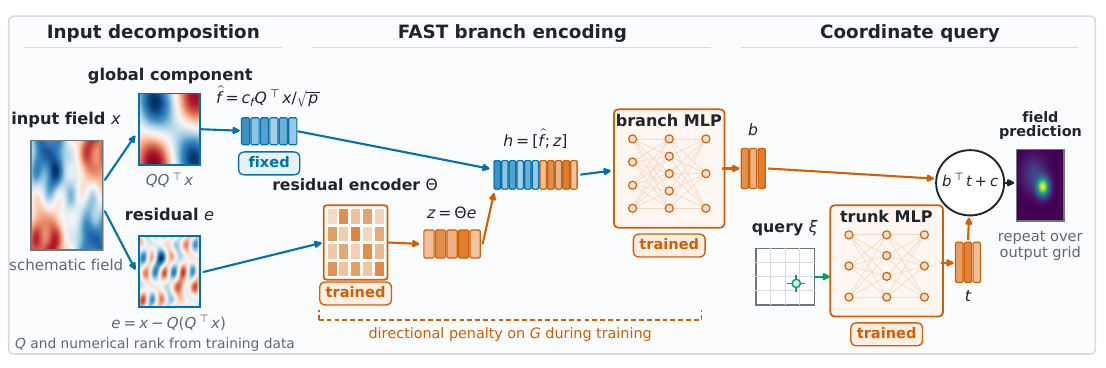}

    \captionof{figure}{
        FAST-DeepONet architecture. A basis estimated from the training
        fields defines a fixed spectral path with scale $c_f$ and an
        orthogonal residual path. The fixed and learned features are
        concatenated and mapped to the branch output $b$, while the trunk
        network maps an output query $\xi$ to $t$. Their inner product
        $b^\top t+c$ gives the predicted field value at that query.
        The diagram uses $x$, $Q$, and $p$ for the field block.
        Standardized auxiliary parameters are appended to the fixed
        representation and excluded from the residual path. The dashed
        bracket denotes the directional penalty used only during training.
    }
    \label{fig:architecture}
\end{minipage}
\par\medskip

% for submission
% \begin{figure}[!htbp]
%     \centering
%     \includegraphics[width=\linewidth]{fast_deeponet_architecture.pdf}
%     \caption{FAST-DeepONet architecture. A basis estimated from the training fields defines a fixed
%         spectral path with scale $c_f$ and an orthogonal residual path. The fixed and learned
%         features are concatenated and mapped to the branch output $b$, while the trunk network maps
%         an output query $\xi$ to $t$. Their inner product $b^\top t+c$ gives the predicted field
%         value at that query. The diagram uses $x$, $Q$, and $p$ for the field block. Standardized
%         auxiliary parameters are appended to the fixed representation and excluded from the residual
%         path. The dashed bracket denotes the directional penalty used only during training.}
%     \label{fig:architecture}
% \end{figure}

\subsection{Spectral and residual branch representation}

The branch input may contain a discretized field $x_f\in\R^{p_f}$ and $p_\eta$ standardized auxiliary physical parameters $\eta_s\in\R^{p_\eta}$.
We write
\begin{equation}
    x=[x_f;\eta_s]\in\R^p,
    \qquad
    p=p_f+p_\eta.
    \label{eq:method_branch_input}
\end{equation}
The spectral decomposition is applied only to the field block. Let $X_f\in\R^{n\times p_f}$ collect the transformed training fields and consider its uncentered thin singular value decomposition
\begin{equation}
    X_f
    =
    U\,\operatorname{diag}(\sigma_1,\sigma_2,\ldots)V^\top.
    \label{eq:training_svd}
\end{equation}

The requested fixed representation has total dimension $r$. Assuming $r>p_\eta$ and $\sigma_1>0$, the requested field dimension is
\begin{equation}
    r_f=r-p_\eta.
\end{equation}
We define the numerical and effective field ranks by
\begin{equation}
    r_{\mathrm{num},f}
    =
    \#\left\{
        i:\sigma_i>10^{-6}\sigma_1
    \right\},
    \qquad
    r_{\mathrm{eff},f}
    =
    \max\left\{
        1,
        \min(r_f,r_{\mathrm{num},f})
    \right\},
    \label{eq:rank_rule_field}
\end{equation}
and set
\begin{equation}
    r_{\mathrm{eff}}
    =
    r_{\mathrm{eff},f}+p_\eta.
    \label{eq:rank_rule}
\end{equation}
The prescribed relative threshold discards numerically unresolved singular directions while retaining at least one field direction.

Let \[Q_f\in\R^{p_f\times r_{\mathrm{eff},f}}[\] contain the leading right singular vectors, so that \[Q_f^\top Q_f=I_{r_{\mathrm{eff},f}}.\]
For a fixed positive scale $c_f$, the spectral field features and orthogonal residual are
\begin{equation}
    \hatf_f(x_f)
    =
    c_f\frac{Q_f^\top x_f}{\sqrt{p_f}},
    \qquad
    e_f(x_f)
    =
    x_f-Q_f(Q_f^\top x_f).
    \label{eq:factor_residual}
\end{equation}
The residual uses the unscaled orthogonal projection and is therefore independent of $c_f$.

The complete fixed representation and residual are
\begin{equation}
    \hatf(x)
    =
    [\hatf_f(x_f);\eta_s],
    \qquad
    e(x)
    =
    [e_f(x_f);\mathbf{0}_{p_\eta}].
    \label{eq:block_preprocessing}
\end{equation}
Thus the auxiliary parameters are passed directly to the fixed representation and excluded from the residual path. In the reported benchmark containing an auxiliary parameter, $c_f=1$. When $p_\eta=0$, the auxiliary blocks are omitted and $r_f=r$.

Define the block residual projector
\begin{equation}
    P_{\mathrm{res}}
    =
    \operatorname{diag}
    \left(
        I_{p_f}-Q_fQ_f^\top,
        \mathbf{0}_{p_\eta\times p_\eta}
    \right).
    \label{eq:block_projector}
\end{equation}
This matrix is symmetric and idempotent, and $e(x)=P_{\mathrm{res}}x$.
The basis $Q_f$ and numerical rank $r_{\mathrm{num},f}$ are computed from the training fields only. The requested total rank and factor scale are fixed before test evaluation. In the implementation, the field residual is evaluated as $x_f-Q_f(Q_f^\top x_f)$, without explicitly forming a dense $p_f\times p_f$ projector.

A trainable matrix
\[
    \Theta\in\R^{s\times p}
\]
maps the residual to
\begin{equation}
    z(x)
    =
    \Theta e(x)
    =
    \Theta P_{\mathrm{res}}x
    \in\R^s.
    \label{eq:residual_features}
\end{equation}
The input to the branch network is
\begin{equation}
    h(x)
    =
    [\hatf(x);z(x)]
    \in\R^{r_{\mathrm{eff}}+s}.
    \label{eq:branch_representation}
\end{equation}
For an output query $\xi$, FAST-DeepONet predicts
\begin{equation}
    \widehat{\calG}(x)(\xi)
    =
    \sum_{\ell=1}^{q}
    b_\ell\bigl(h(x)\bigr)t_\ell(\xi)+c,
    \label{eq:fast_deeponet}
\end{equation}
where $b$ and $t$ denote the branch and trunk outputs.

\subsection{Directional regularization of the effective residual map}

Let $w$ denote the width of the first branch hidden layer. Its weights can be partitioned as
\[
    H_f\in\R^{w\times r_{\mathrm{eff}}},
    \qquad
    H_z\in\R^{w\times s},
\]
where $H_f$ acts on the fixed representation and $H_z$ acts on the learned residual features. The first-layer preactivation is
\begin{equation}
    a^{(1)}(x)
    =
    H_f\hatf(x)+H_z z(x)+\beta^{(1)}
    =
    H_f\hatf(x)+Gx+\beta^{(1)},
    \label{eq:first_branch_layer}
\end{equation}
where
\begin{equation}
    G
    =
    H_z\Theta P_{\mathrm{res}}
    \in\R^{w\times p}
    \label{eq:effective_map}
\end{equation}
is the effective map from the original branch coordinates to the residual contribution in the first hidden layer.

The matrices $H_z$ and $\Theta$ do not separately identify $G$. For any invertible change of coordinates $C\in\R^{s\times s}$ of the learned residual features,
\begin{equation}
    \Theta\mapsto C\Theta,
    \qquad
    H_z\mapsto H_zC^{-1}
    \label{eq:coordinate_gauge}
\end{equation}
leaves $G$, the first-layer preactivation, and the network prediction unchanged. Likewise,
\begin{equation}
    (\Theta+A)P_{\mathrm{res}}
    =
    \Theta P_{\mathrm{res}}
\end{equation}
for any $A$ satisfying $AP_{\mathrm{res}}=0$. 
A penalty applied directly to $\Theta$ would therefore depend on an arbitrary factorization and on components that do not affect the prediction.

For each row of $G$, define the exact normalization
\begin{equation}
    \widetilde G_{i,:}
    =
    \begin{cases}
        \displaystyle
        \frac{G_{i,:}}{\lVert G_{i,:}\rVert_2},
        &
        \lVert G_{i,:}\rVert_2>0,
        \\[8pt]
        \mathbf{0}^\top,
        &
        \lVert G_{i,:}\rVert_2=0.
    \end{cases}
    \label{eq:row_normalization}
\end{equation}
No additive constant is used in the denominator, and an exactly zero row remains zero. The directional clipped penalty is
\begin{equation}
    \overline\Omega(G;\tau)
    =
    \frac{1}{wp}
    \sum_{i=1}^{w}
    \sum_{j=1}^{p}
    \min
    \left\{
        \frac{|\widetilde G_{ij}|}{\tau},
        1
    \right\}.
    \label{eq:directional_penalty}
\end{equation}

Consider a positive diagonal matrix $D=\operatorname{diag}(d_1,\ldots,d_w)$. Multiplying the incoming weights and biases of the first-layer ReLU units by $D$ and the corresponding columns of the next-layer weight matrix by $D^{-1}$ leaves the network function unchanged. Under this transformation, $G$ is replaced by $DG$. Since $d_i>0$,
\begin{equation}
    \widetilde{DG}=\widetilde G,
\end{equation}
including for exactly zero rows.
The penalty is therefore invariant to positive rescalings of the first-layer ReLU units, the symmetry behind path-normalized optimization \citep{neyshabur2015pathsgd}.
Together with \eqref{eq:coordinate_gauge}, the regularizer depends on the row directions of the realized effective residual map rather than on a particular factorization or scaling of adjacent trainable layers.

Exact normalization is what buys that invariance, and it costs continuity at the origin. For a row $g=\epsilon e_j$ the normalized row is $e_j$ for every $\epsilon>0$, so the row contributes $1/(wp)$ in the limit, whereas \eqref{eq:row_normalization} assigns a zero row the value zero. The penalty is therefore discontinuous on the set of zero rows and is not differentiable there. We treat \eqref{eq:directional_penalty} as an objective on rows of nonzero norm; the zero-row convention only keeps the map well defined. Replacing the denominator by $\lVert G_{i,:}\rVert_2+\epsilon$ would restore continuity but destroy the exact invariance above, which is the property the construction exists to provide. The set is never approached in practice: across every reported fit no row of $G$ is exactly zero and the smallest row norm is $0.265$, as Appendix~\ref{app:map_diagnostics} reports.

Let $\calD_{\mathrm{train}}$ contain the sampled scalar query pairs from the training operator samples. The training objective is
\begin{equation}
    \frac{1}{|\calD_{\mathrm{train}}|}
    \sum_{(i,j)\in\calD_{\mathrm{train}}}
    \left[
        y_{ij}
        -
        \widehat{\calG}(x_i)(\xi_{ij})
    \right]^2
    +
    \lambda\overline\Omega(G;\tau_k).
    \label{eq:training_objective}
\end{equation}
For epochs $k=1,\ldots,N_{\mathrm{ep}}$, the clipping threshold follows
\begin{equation}
    \tau_k
    =
    \max
    \left\{
        0.01,
        0.1-\frac{0.09k}{N_{\mathrm{ep}}}
    \right\}.
    \label{eq:tau_schedule}
\end{equation}

\subsection{Training procedure}

Algorithm~\ref{alg:training} summarizes one FAST-DeepONet fit. Test fields are not used for preprocessing, optimization, or checkpoint selection.

\begin{algorithm}[!ht]
\caption{FAST-DeepONet training}
\label{alg:training}
\begin{algorithmic}[1]
\Require training and tuning operator samples
$\calD_{\mathrm{tr}},\calD_{\mathrm{tu}}$
\Require fixed $r,c_f,s,\lambda$ and number of epochs $N_{\mathrm{ep}}$
\State estimate input and target transformations from $\calD_{\mathrm{tr}}$
\State apply the transformations to $\calD_{\mathrm{tr}}$ and $\calD_{\mathrm{tu}}$
\State compute $Q_f$ and $r_{\mathrm{eff}}$ from $\calD_{\mathrm{tr}}$ using \eqref{eq:rank_rule_field} and \eqref{eq:rank_rule}
\State initialize $\Theta$, the branch network, and the trunk network
\For{$k=1,\ldots,N_{\mathrm{ep}}$}
    \State draw the prescribed balanced query subset from each training field
    \State set
    $\tau_k\gets
    \max\{0.01,0.1-0.09k/N_{\mathrm{ep}}\}$
    \State form
    $h(x)=[\hatf(x);\Theta P_{\mathrm{res}}x]$
    \State update the trainable parameters using \eqref{eq:training_objective}
    \State retain the checkpoint if the tuning MSE on $\calD_{\mathrm{tu}}$ decreases
\EndFor
\State \Return retained FAST-DeepONet model
\end{algorithmic}
\end{algorithm}

\section{Numerical experiments}
\label{sec:experiments}

\subsection{PDE benchmarks}
\label{subsec:benchmarks}

We consider three controlled operator learning problems: Navier--Stokes flow, Darcy flow, and signed terminal wave prediction. The governing equations are standard and the input distributions are defined for this study. All input sensors have fixed locations. For Darcy and wave, the candidate disk locations are also fixed across the training, tuning, and test sets, while the active subset and its amplitudes vary by sample. The two benchmarks use different candidate site sets.

\paragraph{Navier--Stokes flow.}
We solve the incompressible Navier--Stokes equations on the periodic unit square,
\begin{equation}
    \partial_t\boldsymbol v+(\boldsymbol v\cdot\nabla)\boldsymbol v
    =-\nabla p+\nu\Delta\boldsymbol v,
    \qquad \nabla\cdot\boldsymbol v=0,
    \label{eq:nse}
\end{equation}
using a pseudo-spectral solver with $2/3$ dealiasing, pressure projection, explicit advection, and implicit treatment of diffusion. The solver uses a $128\times128$ grid, $\Delta t=10^{-3}$, and $T=0.2$. The initial velocity is
\begin{align}
    v_{1,0}(x,y)&=-a_1\sin(2\pi y+\phi_y)
      +a_3\sin(2\pi x)\sin(2\pi y),\nonumber\\
    v_{2,0}(x,y)&= a_2\sin(4\pi x+\phi_x)
      +a_3\cos(2\pi x)\cos(2\pi y),
    \label{eq:nse_initial}
\end{align}
which is divergence free.
Here $\phi_x,\phi_y\sim\mathcal U(0,2\pi)$,
$a_1,a_2\sim\mathcal U(0.8,1.2)$,
$a_3\sim\mathcal U(-0.25,0.25)$, and
$\nu\sim\mathcal U(0.01,0.05)$ independently.
The branch contains both velocity components sampled on a $64\times64$ grid and the viscosity, giving $p=2(64^2)+1=8193$. We use the terminal horizontal velocity $v_1(\cdot,T)$ as the physical target. During training, both models predict its increment relative to a bilinear interpolation of the initial horizontal velocity; the full terminal field is reconstructed before evaluation. This common residual target removes the nearly identity part of the evolution.

\paragraph{Darcy flow.}
We consider the Darcy map from permeability to pressure head used widely in operator learning \citep{bhattacharya2021model,li2021fourier},
\begin{equation}
    -\nabla\cdot\left(a(x)\nabla u(x)\right)=1,\qquad
    u|_{\partial(0,1)^2}=0,
    \label{eq:darcy}
\end{equation}
on a $64\times64$ grid.
The discretization is a five point finite difference scheme. Because the coefficient varies in space, the face transmissibilities are harmonic means of the two adjacent cell values. The branch input is the log permeability $g=\log a$ with background
\begin{equation}
    g_{\mathrm{bg}}(x)
    =0.8\sum_{k=1}^{8} f_k e^{-(k-1)/4}\phi_k(x),
    \qquad f_k\sim\mathcal N(0,1),
    \label{eq:darcy_input}
\end{equation}
where $\{\phi_k\}$ are the separable Fourier modes listed in Appendix~\ref{app:protocol}. Each sample activates $3$--$6$ of ten candidate disks. Every disk has radius $0.045$, an independently sampled sign, and magnitude drawn uniformly from $[1.2,1.8]$. The complete field is clipped to $[-3,3]$ before setting $a=e^g$. The branch and target are $g$ and the pressure head $u$, respectively, both on the native grid, and $p=4096$.

\paragraph{Signed terminal wavefield.}
On the periodic unit square, the heterogeneous wave equation is
\begin{equation}
    m(x)\,\partial_{tt}u(x,t)=\Delta u(x,t)
    \label{eq:wave}
\end{equation}
with squared slowness
\begin{equation}
    m(x)=1+\frac{0.35}{\sqrt 6}\sum_{k=1}^{6}f_k\phi_k(x)
    +\sum_{\ell\in A}\alpha_\ell\mathbf 1_{D_\ell}(x).
    \label{eq:wave_input}
\end{equation}
The coefficients satisfy $f_k\sim\mathcal U(-1,1)$. The active set $A$ contains $5$--$8$ of ten candidate disks, each with radius $0.05$ and $\alpha_\ell\sim\mathcal U(0.5,0.8)$. The initial displacement is a Gaussian centered at $(0.5,0.5)$ with standard deviation $0.05$, and the initial velocity is zero. A second-order leapfrog method with the corresponding zero-velocity initialization advances the solution on a $64\times64$ grid using $\Delta t=0.004$ to $T=0.4$. The branch input is $m-1$, so $p=4096$, and the target is the signed terminal field $u(\cdot,T)$. Appendix~\ref{app:wave_propagation} shows the propagation from the localized pulse to the terminal target, while Appendix~\ref{app:checks} reports CFL and refinement checks.

Figure~\ref{fig:benchmarks} shows one input and output pair from each test set, with every panel labeled by the variable it displays. The Navier--Stokes panel shows the initial horizontal velocity alone, and the wave panel shows $m$ rather than the stored $m-1$.

\par\medskip
\noindent
\begin{minipage}{\linewidth}
    \centering
    \includegraphics[width=0.8\linewidth]
        {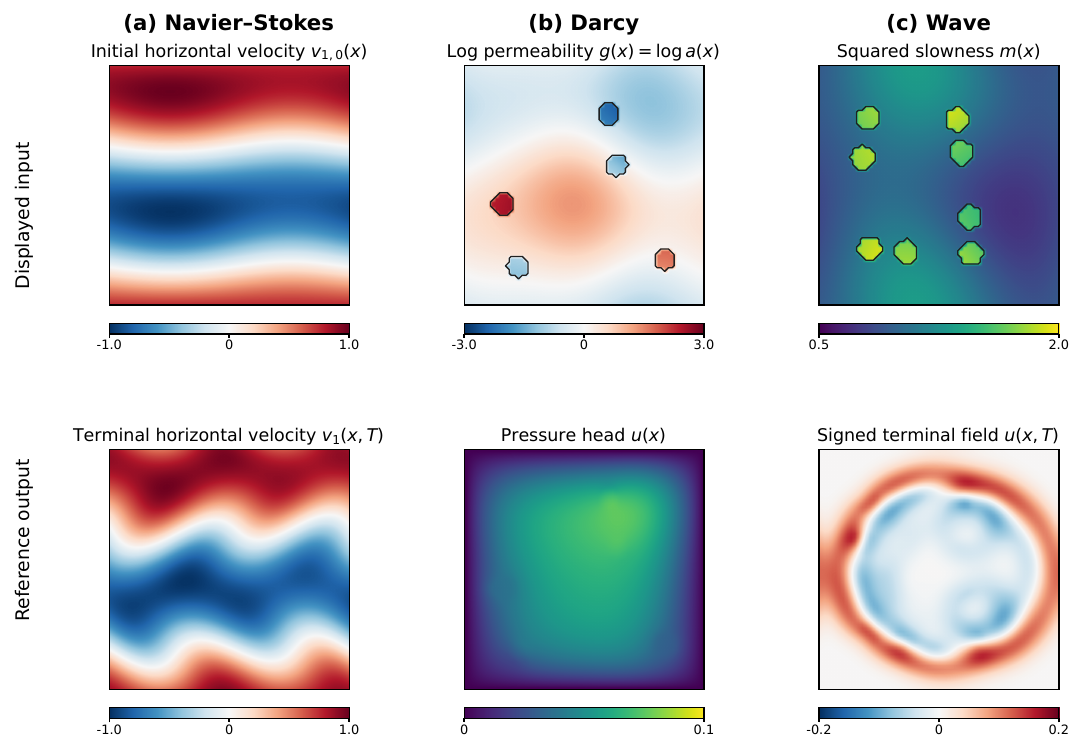}

    \captionof{figure}{
        Representative inputs and outputs, selected without using model
        predictions or errors. Black contours mark the active disks in the two
        designed input distributions and are not supplied to the models. Each
        colorbar applies only to its panel.
    }
    \label{fig:benchmarks}
\end{minipage}
\par\medskip

% for submission
% \begin{figure}[!htbp]
%     \centering
%     \includegraphics[width=0.8\linewidth]{figure2_benchmark_inputs_targets.pdf}
%     \caption{Representative inputs and outputs, selected without using model predictions or errors.
%         Black contours mark the active disks in the two designed input distributions and are not
%         supplied to the models. Each colorbar applies only to its panel.}
%     \label{fig:benchmarks}
% \end{figure}

\subsection{Experimental protocol}
\label{subsec:protocol}

Plain DeepONet and FAST-DeepONet use the same data, output queries, target transformations, branch and trunk depths, hidden widths, optimizer, and checkpoint rule. They differ only in the representation supplied to the branch. Both networks have three hidden layers, branch and trunk width $128$, latent dimension $128$, and ReLU activations. Dropout is not used for the point prediction experiments.

Adam starts at $10^{-3}$, and an exponential scheduler multiplies the learning rate by $0.995$ after each epoch. Each fit runs for $300$ epochs with $8192$ balanced scalar queries per epoch, batch size $2048$, and a fixed set of $4096$ tuning queries. The checkpoint with the lowest tuning mean squared error is retained. That optimum lies inside the budget for every reported fit, and Appendix~\ref{app:training_curves} shows the curves. Navier--Stokes and wave use periodic Fourier coordinate features through order four in the trunk. Darcy uses the raw coordinates together with the same Fourier features. Targets are divided by one scalar standard deviation estimated from the training fields. Darcy and wave targets are first centered by their pointwise training mean; the Navier--Stokes increment is not centered. All transformations are reversed before evaluation, which uses every point on the native output grid.

\begin{table}[!ht]
\centering
\caption{Final benchmark and FAST representation settings. The rank column gives the requested and effective factor ranks. For Navier--Stokes, the effective representation contains five field directions and one separately standardized viscosity scalar. The wave training set is a fixed subset of a fresh $60$ field pool.}
\label{tab:protocol}
\setlength{\tabcolsep}{4pt}
\begin{tabular}{lrrrrrr}
\toprule
Benchmark & $n_{\mathrm{tr}}$ & $p$ & Output grid & Rank & $s$ & $(c_f,\lambda)$\\
\midrule
Navier--Stokes & 60 & 8193 & $128^2$ & $8/6$ & 8 & $(1,0.01)$\\
Darcy & 60 & 4096 & $64^2$ & $16/16$ & 16 & $(1,1)$\\
Terminal wave & 40 & 4096 & $64^2$ & $6/6$ & 16 & $(8,1)$\\
\bottomrule
\end{tabular}
\end{table}

Navier--Stokes and Darcy use $60$ training fields; wave uses the fixed $40$ field subset in Table~\ref{tab:protocol}. Every benchmark has $10$ tuning fields and $100$ disjoint test fields. Model settings were fixed before the final test set was generated. The three studies were frozen independently and share a common evaluation protocol. Appendix~\ref{app:protocol} gives the split sizes, run settings, input spectra, generator details, and how each configuration was fixed. Appendix~\ref{app:fixed_budget} repeats one paired seed per benchmark under a common $200$ epoch cap, and Appendix~\ref{app:cost} reports parameter counts and training cost.

Within a benchmark, Plain and FAST share the data and query schedule. Tables report the mean and sample standard deviation of the test mean across these paired fits. Two intervals are reported for their difference: a paired Student $t$ interval across fits, and a percentile bootstrap that resamples fit and field indices together, keeping the models paired. The fieldwise error is
\begin{equation}
    E_i
    =\frac{\lVert y_i-\widehat y_i\rVert_2}
           {\lVert y_i\rVert_2+10^{-12}},
    \label{eq:relative_l2}
\end{equation}
where $y_i$ and $\widehat y_i$ denote the full target and prediction for field $i$.

\subsection{Point prediction accuracy}
\label{subsec:main_results}

Table~\ref{tab:main} and Figure~\ref{fig:accuracy} report the main comparison on the independent test sets. FAST-DeepONet reduces mean relative $L_2$ error by $37.0\%$ for Navier--Stokes, $10.1\%$ for Darcy, and $4.7\%$ for the signed terminal wavefield. Both intervals in Table~\ref{tab:contrasts} exclude zero for each benchmark.

\begin{table}[!ht]
\centering
\caption{Mean fieldwise relative $L_2$ error on $100$ independent test fields.}
\label{tab:main}
\setlength{\tabcolsep}{7pt}
\begin{tabular}{lccc}
\toprule
Branch representation & Navier--Stokes & Darcy & Terminal wave\\
\midrule
Plain DeepONet & $0.1776\pm0.0057$ & $0.1029\pm0.0022$ & $0.2165\pm0.0030$\\
FAST-DeepONet & $\mathbf{0.1119\pm0.0021}$ & $\mathbf{0.0925\pm0.0019}$ & $\mathbf{0.2063\pm0.0026}$\\
\bottomrule
\end{tabular}
\end{table}

\par\medskip
\noindent
\begin{minipage}{\linewidth}
    \centering
    \includegraphics[width=\linewidth]
        {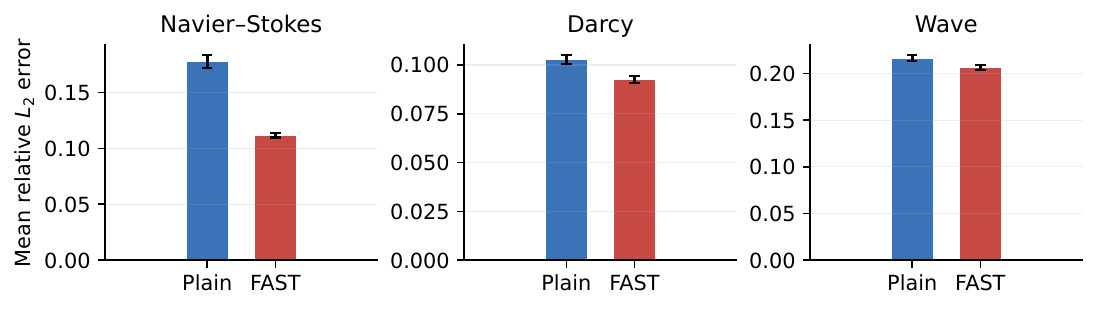}
    \captionof{figure}{Plain and FAST test errors for the three PDE problems, where lower is better. Bars and whiskers show the mean and sample standard deviation across paired fits. Exact values are given in Table~\ref{tab:main}.}
    \label{fig:accuracy}
\end{minipage}
\par\medskip

% for submission
% \begin{figure}[!htbp]
%     \centering
%     \includegraphics[width=\linewidth]{figure3_plain_fast_accuracy.pdf}
%     \caption{Plain and FAST test errors for the three PDE problems, where lower is better. Bars and
%         whiskers show the mean and sample standard deviation across paired fits. Exact values are
%         given in Table~\ref{tab:main}.}
%     \label{fig:accuracy}
% \end{figure}

\begin{table}[!ht]
\centering
\caption{Paired FAST minus Plain differences in mean relative $L_2$ error. Negative values favor FAST. The bootstrap resamples fit and field indices while retaining model pairing.}
\label{tab:contrasts}
\setlength{\tabcolsep}{6pt}
\begin{tabular}{lrrr}
\toprule
Benchmark & Mean difference & Paired fit $95\%$ CI & Two-way bootstrap $95\%$ CI\\
\midrule
Navier--Stokes & $-0.0657$ & $[-0.0737,-0.0578]$ & $[-0.0816,-0.0511]$\\
Darcy & $-0.0104$ & $[-0.0143,-0.0064]$ & $[-0.0159,-0.0048]$\\
Terminal wave & $-0.0103$ & $[-0.0160,-0.0046]$ & $[-0.0164,-0.0042]$\\
\bottomrule
\end{tabular}
\end{table}

Figure~\ref{fig:qualitative_fields} complements the aggregate errors with one example from each benchmark, again selected without using predictions or errors. Ground truth, Plain, and FAST share a solution scale within each benchmark, and the two error maps share a second scale.

\par\medskip
\noindent
\begin{minipage}{\linewidth}
    \centering
    \includegraphics[width=\linewidth]
        {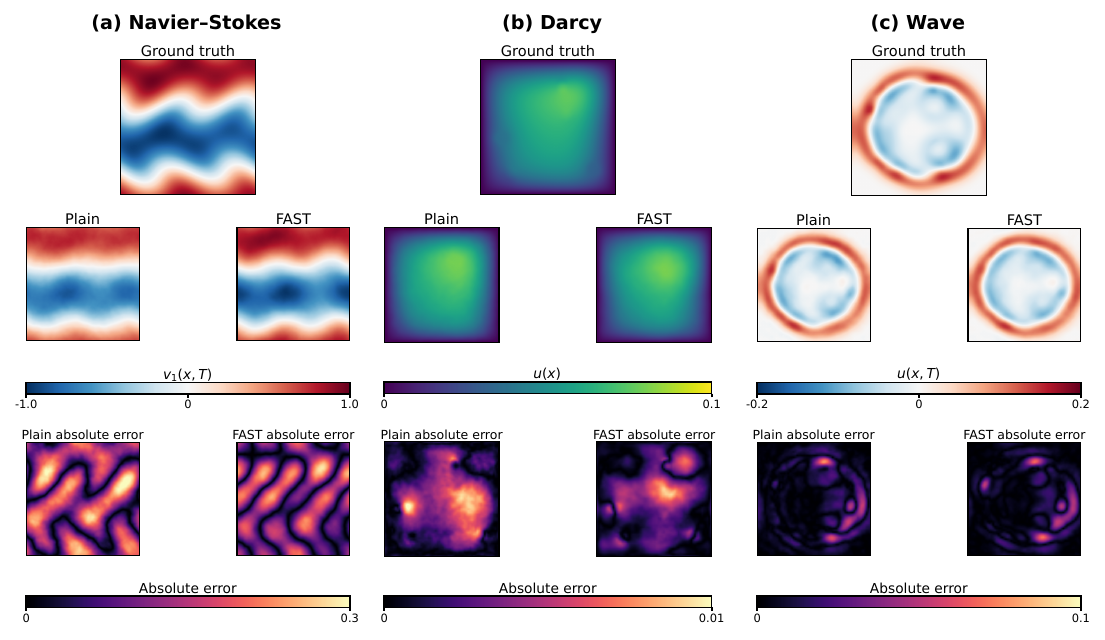}
    \captionof{figure}{ Representative Plain and FAST reconstructions under the main training protocol. Each block shows the ground truth, both predictions, and absolute error maps, in which smaller values are better.}
    \label{fig:qualitative_fields}
\end{minipage}
\par\medskip

% for submission
% \begin{figure}[!htbp]
%     \centering
%     \includegraphics[width=\linewidth]{figure4_field_comparison.pdf}
%     \caption{Representative Plain and FAST reconstructions under the main training protocol. Each block
%         shows the ground truth, both predictions, and absolute error maps, in which smaller values
%         are better.}
%     \label{fig:qualitative_fields}
% \end{figure}

\subsection{Dependence on input dimension and sample size}
\label{subsec:pn}

The main comparisons use the high-dimensional, small-sample setting that motivates the branch representation. We therefore vary the Navier--Stokes input grid over $q\in\{8,16,32,64\}$, giving $p=2q^2+1\in\{129,513,2049,8193\}$, and use nested training sets with $n\in\{20,60,240\}$. The sweep draws its nested subsets from a separate bank of $240$ Navier--Stokes fields, so its cells are not numerically comparable with Table~\ref{tab:main}. The target and output grid remain fixed at $128\times128$. The number of scalar queries per epoch is scaled with $n$ to keep the average query budget per training field approximately constant. Changing $q$ moves both the branch dimension and the resolved input information, so this is a joint sensor resolution and dimension diagnostic. The same diagnostic is repeated on Darcy and terminal wave over $q\in\{16,32,64\}$, where the branch holds a single field and $p=q^2$, each on a fresh training bank constructed in Appendix~\ref{app:pn_full}.

At $n=240$, Plain achieves mean error $0.0394$ for $p=129$ but rises to $0.1556$ for $p=8193$. FAST changes from $0.0422$ to $0.0408$ over the same range. Thus Plain performs normally in the low-dimensional, data-rich corner, while FAST remains stable as the sensor grid grows. At $n=20$, both methods remain near $0.2$, showing that the representation does not compensate for severe sample scarcity. The two branches also differ in how much one fit depends on the training randomness. The spread of the test mean across the paired seeds is narrower for FAST at three of the four corners, by factors of $4.4$ and $9.9$ at $p=8193$ and $15.1$ at $p=129$ with $n=20$. The largest of these occurs where the accuracy difference is not resolved, so reproducibility separates the two branches in a regime where the mean error does not. The exception is $p=129$ with $n=240$, where a plain first layer is already well determined and the plain branch is the more reproducible of the two. Appendix~\ref{app:pn_full} reports those four corners with their paired fits.

Figure~\ref{fig:pn_sweep_all} places the three benchmarks on a common axis. At $n=240$ the reduction against a plain branch grows with the sensor grid on all three, from $8.6\%$ to $73.8\%$ on Navier--Stokes, from $16.0\%$ to $31.1\%$ on Darcy, and from $-34.3\%$ to $5.6\%$ on terminal wave. The wave curve stays below the other two and turns positive only at the full grid. Appendix~\ref{app:pn_full} gives every fitted cell.

\par\medskip
\noindent
\begin{minipage}{\linewidth}
    \centering
    \includegraphics[width=\linewidth]{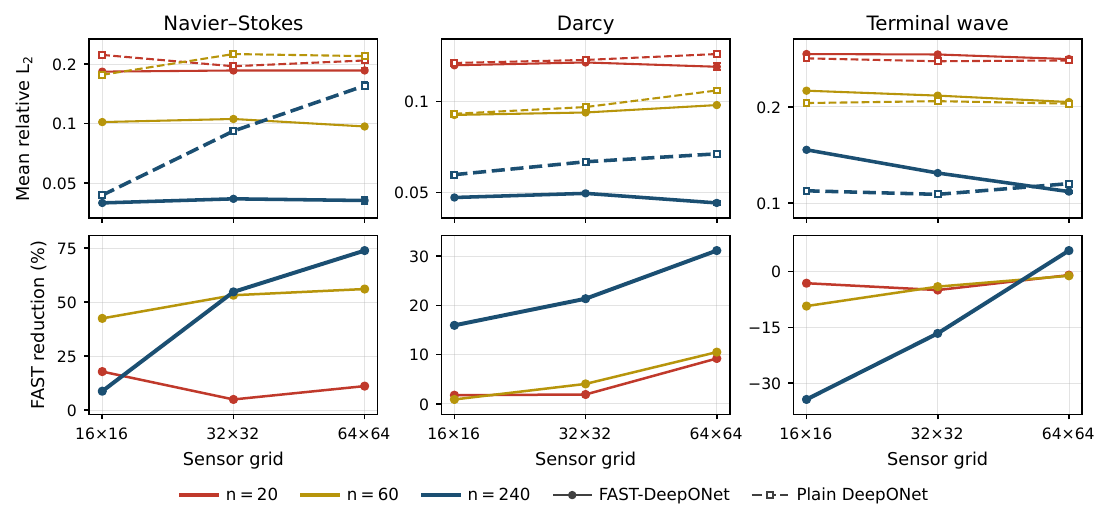}
    \captionof{figure}{Sensor resolution and sample size on the three benchmarks. Top: mean test error, lower is better. Bottom: the FAST reduction against a plain branch in the same cell, higher is better, so a rising curve means the advantage grows with the branch dimension. Corner cells carry five paired seeds and a one standard deviation whisker; interior cells carry one seed. The designed benchmarks start at $16\times16$; Appendix~\ref{app:pn_full} explains why and Table~\ref{tab:app_pn_sweep} keeps the $8\times8$ row.}
    \label{fig:pn_sweep_all}
\end{minipage}
\par\medskip

% for submission
% \begin{figure}[!htbp]
%     \centering
%     \includegraphics[width=\linewidth]{pn_sweep_all.pdf}
%     \caption{Sensor resolution and sample size on the three benchmarks. Top: mean test error, lower is
%         better. Bottom: the FAST reduction against a plain branch in the same cell, higher is
%         better, so a rising curve means the advantage grows with the branch dimension. Corner cells
%         carry five paired seeds and a one standard deviation whisker; interior cells carry one seed.
%         The designed benchmarks start at $16\times16$; Appendix~\ref{app:pn_full} explains why and
%         Table~\ref{tab:app_pn_sweep} keeps the $8\times8$ row.}
%     \label{fig:pn_sweep_all}
% \end{figure}

\subsection{Contribution of the spectral and residual paths}
\label{subsec:mechanism}

To separate the two paths we fit a spectral-only branch. It receives the same basis, effective rank, factor scale, downstream widths, data, query schedule, and model seeds as FAST-DeepONet, and omits only the residual route. This is the input-side reduced-basis reference: the branch is compressed with a fixed spectral basis estimated from the training inputs, so any remaining difference isolates the residual path.

The residual path contributes differently on each benchmark. For Navier--Stokes the reference reaches $0.1108$ against $0.1119$ for FAST and the paired interval includes zero. For Darcy it reaches $0.0876$ against $0.0925$, so the residual path costs a small amount in that configuration. For terminal wave prediction FAST lowers the reference error from $0.2557$ to $0.2063$, a $19.3\%$ reduction. The fixed spectral path therefore accounts for essentially all of the improvement over a plain branch on two of the three benchmarks, and the question is what distinguishes the third.

The retained rank is the relevant difference. We form the field residual $e_f(x_f)$ of \eqref{eq:factor_residual} and compare its per-coordinate variance with that of the input. At the Darcy rank of $16$ the spectral path has already absorbed all but $3.8\%$ of the input variance on the scatterer coordinates, leaving little for the residual path; at the wave rank of $6$ the residual still retains $61.5\%$. Across these three benchmarks the residual path contributes where the retained spectral subspace leaves a substantial part of the input variation unexplained, and the spectral path is sufficient where it does not. Appendix~\ref{app:map_diagnostics} reports the full diagnostics and characterizes the learned map.

The directional penalty is what makes that contribution possible. Refitting with $\lambda=0$ and changing nothing else, the unregularized residual path is indistinguishable from the spectral-only reference on terminal wave and is actively harmful on Darcy. The improvement therefore comes from regularizing the map, not from adding the path. Appendix~\ref{app:ablation} reports the full paired contrasts and Appendix~\ref{app:penalty} the penalty ablation. Appendix~\ref{app:controls} adds two controls: a random basis of the same rank, which loses to a plain branch on all three benchmarks, and a plain branch narrowed to the FAST parameter count, which is worse than the full-width plain branch. Appendix~\ref{app:scaling} reports the sensitivity to the factor scale. Appendix~\ref{app:recipe} collects what those studies suggest for choosing the representation parameters on a new problem.

The point prediction results above use deterministic networks. Appendix~\ref{app:conformal} reports a separate study in which Plain and FAST are retrained with dropout and calibrated on fresh data for operator-level uncertainty quantification.

\section{Conclusion}
\label{sec:conclusion}

FAST-DeepONet combines fixed spectral factors with a regularized projection of the orthogonal residual to stabilize high-dimensional DeepONet branches. The regularizer acts on the normalized effective residual map, which removes the dependence on nonidentifiable parameter scalings. The resulting model trains three to seven times fewer parameters than a plain branch.

Across the three benchmarks it lowers the error of a plain branch by $4.7\%$ to $37.0\%$. A separate sweep over the sensor grid shows how that margin depends on the regime. On all three benchmarks it grows with the branch dimension, reaching nearly a factor of four on Navier--Stokes at $p=8193$, and it disappears when the branch is small. Neither representation compensates for severe sample scarcity. The two paths do not contribute equally. The fixed spectral path carries the gain on Navier--Stokes and Darcy, while terminal wave prediction requires the residual path together with its directional penalty. Without either one the representation falls behind a plain branch on that benchmark.

These results cover three generated PDE families with input distributions designed for this study, fixed sensor layouts, and $40$ to $60$ training operators. Future work will consider varying sensor layouts, broader PDE distributions, and adaptive residual capacity.

\section*{Acknowledgement}
We would like to thank the support of National Science Foundation (DMS-2533878, DMS-2053746, DMS-2134209, ECCS-2328241, CBET-2347401 and OAC-2311848), and U.S.~Department of Energy (DOE) Office of Science Advanced Scientific Computing Research program under the "Uncertainty Quantification for Multifidelity Operator Learning (MOLUcQ)" project (Project No. 81739), DE-SC0023161, the SciDAC LEADS Institute, and DOE–Fusion Energy Science, under grant number: DE-SC0024583.

\section*{CRediT authorship contribution statement}

\textbf{Jiyong Kwon}: Conceptualization, Methodology, Software, Formal analysis, Investigation, Writing -- original draft.
\textbf{BongSeok Kim}: Data curation, Methodology, Investigation, Writing -- review and editing.
\textbf{Guang Lin}: Supervision, Conceptualization, Writing -- review and editing.

\section*{Declaration of competing interest}

The authors declare that they have no known competing financial interests or personal relationships that could have appeared to influence the work reported in this paper.

\section*{Code and data availability}

The code, PDE generators, configurations, and scripts used to reproduce the reported results are available at \url{https://github.com/kwonji39/fast-deeponet}.

\section*{Declaration of generative AI and AI-assisted technologies in the manuscript preparation process}

During the preparation of this work, the authors used Anthropic Claude for language editing and assistance with code. No generative image model was used; After using these tools, the authors reviewed and verified the resulting text, code, and numerical results and take full responsibility for the content of the article.

\appendix
\numberwithin{equation}{section}
\numberwithin{table}{section}
\numberwithin{figure}{section}
\renewcommand*{\theHequation}{\thesection.\arabic{equation}}
\renewcommand*{\theHtable}{\thesection.\arabic{table}}
\renewcommand*{\theHfigure}{\thesection.\arabic{figure}}

\section{Data separation and reproducibility}
\label{app:protocol}

Each configuration was fixed using development data before its independent final accuracy stream was opened for evaluation. The aligned table combines three independently frozen studies. Training and tuning fields may be used to estimate preprocessing, choose configurations, and retain checkpoints, while accuracy test fields are evaluation only. For the conformal study, the dropout architecture, score, levels, Monte Carlo rule, and fitting protocol were frozen before generating a further calibration stream and a disjoint UQ test stream. Neither stream enters model fitting or configuration selection.

The way each benchmark was fixed should be stated plainly. Within a benchmark, the data regime and the number of training fields were chosen on development data. The rule favored the largest FAST reduction among the configurations that passed predeclared validity gates. The FAST representation parameters were then chosen by mean tuning MSE across a development seed schedule. Plain DeepONet has no corresponding parameters. Every reported number comes from a stream generated after those choices were frozen, so the selection governs which regimes are shown rather than the outcome within a regime. Appendices~\ref{app:penalty} and \ref{app:scaling} quantify what the FAST-specific settings are worth once a regime is fixed.

Table~\ref{tab:app_settings} collects the run settings that are common to the three benchmarks together with the entries that differ between them. It complements Table~\ref{tab:protocol}, which lists the branch dimensions and the FAST representation parameters.

\begin{center}
\begin{minipage}{\linewidth}
\centering
\captionof{table}{Run settings for the final study. Entries spanning the three columns are shared by all benchmarks. Wave uses a fixed subset of a fresh training pool. Conformal calibration and UQ streams are independent of the accuracy test streams.}
\label{tab:app_settings}
\small
\setlength{\tabcolsep}{4pt}
\begin{tabular}{lccc}
\toprule
Setting & Navier--Stokes & Darcy & Terminal wave\\
\midrule
\multicolumn{4}{l}{\emph{Data}}\\
Training fields & 60 & 60 & 40 of 60\\
Tuning fields & \multicolumn{3}{c}{10}\\
Accuracy test fields & \multicolumn{3}{c}{100}\\
UQ calibration and test fields & \multicolumn{3}{c}{20 and 100}\\
\addlinespace[3pt]
\multicolumn{4}{l}{\emph{Targets and queries}}\\
Prediction target & increment & pressure head & signed terminal field\\
Pointwise mean centering & none & training mean & training mean\\
Trunk coordinates & Fourier & raw and Fourier & Fourier\\
Fourier order & \multicolumn{3}{c}{4}\\
Target scaling & \multicolumn{3}{c}{one training standard deviation}\\
\addlinespace[3pt]
\multicolumn{4}{l}{\emph{Optimization}}\\
Epochs, main study & \multicolumn{3}{c}{300}\\
Epochs, budget study & \multicolumn{3}{c}{200}\\
Queries per epoch & \multicolumn{3}{c}{8192 training, 4096 tuning}\\
Batch size & \multicolumn{3}{c}{2048}\\
Optimizer & \multicolumn{3}{c}{Adam, $10^{-3}$, decay $0.995$ per epoch}\\
Checkpoint & \multicolumn{3}{c}{lowest tuning MSE}\\
Model seeds & \multicolumn{3}{c}{5, shared by all representations}\\
Representations & \multicolumn{3}{c}{Plain, spectral only, FAST}\\
\bottomrule
\end{tabular}
\end{minipage}
\end{center}

The spectral-only reference shares the same seed vector, so all three representations are paired. Seed values are recorded in the released result archives, which also store the test paths, targets, fieldwise errors, predictions, selected checkpoint, and preprocessing metadata. An audit verified identical test paths and targets across the paired representations. The Navier--Stokes spectral-only fit was added after the Plain and FAST test evaluation. Its architecture, preprocessing, paired seeds, and query schedule were nevertheless frozen before those fits, and none of them was selected using the test outcomes.

The basis functions used in \eqref{eq:darcy_input} and \eqref{eq:wave_input} are, in this order,
\begin{equation*}
\begin{aligned}
&\cos(2\pi x), && \cos(2\pi y), && \cos(2\pi x)\cos(2\pi y), && -\sin(2\pi x),\\
&-\sin(2\pi y), && \sin(2\pi x)\sin(2\pi y), && \cos(4\pi x), && \cos(4\pi y),
\end{aligned}
\end{equation*}
and wave uses the first six. For each heterogeneous benchmark, the ten candidate sites are generated once by selecting ten points from the $4\times4$ grid $\{0.2,0.4,0.6,0.8\}^2$ and adding independent uniform jitter in $[-0.04,0.04]^2$. This placement is drawn once per benchmark, so the two benchmarks use different site sets and each site set remains fixed across all splits within its benchmark.

Table~\ref{tab:app_spectra} gives training input diagnostics for the final datasets. Numerical rank uses the uncentered singular values and the tolerance in \eqref{eq:rank_rule_field}. Effective rank and $k_{95}$ use the centered covariance, where effective rank is the exponential of the eigenvalue entropy and $k_{95}$ is the number of components required to explain $95\%$ of the variance. Each diagnostic is computed on the branch variable actually supplied to the model, so the wave row refers to $m-1$. For Navier--Stokes that variable also carries the viscosity, and its numerical rank of $6$ counts five field directions together with the auxiliary coordinate. The rank rule in \eqref{eq:rank_rule_field} sees only the five, which with $p_\eta=1$ gives the effective rank of $6$ in Table~\ref{tab:protocol}.

\begin{center}
\begin{minipage}{\linewidth}
\centering
\captionof{table}{Training input spectral diagnostics for the final datasets. These values describe the data and are distinct from the selected factor ranks in Table~\ref{tab:protocol}.}
\label{tab:app_spectra}
\setlength{\tabcolsep}{7pt}
\begin{tabular}{lrrr}
\toprule
Benchmark & Numerical rank & Effective rank & $k_{95}$\\
\midrule
Navier--Stokes & 6 & 4.05 & 4\\
Darcy & 31 & 5.05 & 9\\
Terminal wave & 16 & 10.76 & 12\\
\bottomrule
\end{tabular}
\end{minipage}
\end{center}

The qualitative examples in Figure~\ref{fig:qualitative_fields} use the first paired model seed and prediction-independent field rules. Navier--Stokes uses field 95, selected by a target-only dynamics score; Darcy uses the prespecified first test field; and terminal wave uses field 14, selected by a target-only anomaly score. They are presentation examples and enter no quantitative result.

\section{Spectral-only reference}
\label{app:ablation}

Section~\ref{subsec:mechanism} defines the spectral-only branch. Table~\ref{tab:app_ablation} reports the full comparison.

The reference is a principal component reduction of the branch input, the basis that non-intrusive reduced order models pair with a network \citep{hesthaven2018nonintrusive}. By \eqref{eq:training_svd} the columns of $Q_f$ are the leading right singular vectors of the uncentered training input matrix. The spectral path therefore spans the subspace of an uncentered principal component analysis, and its features differ from the corresponding scores only by the fixed factor $c_f/\sqrt{p_f}$. A centered analysis of the same inputs is a different reduction, at a maximum principal angle of $13.6$ degrees; we use the uncentered form because the rank safeguard in \eqref{eq:rank_rule_field} is defined on the uncentered singular values.

\begin{center}
\begin{minipage}{\linewidth}
\centering
\captionof{table}{Spectral-only reference on the independent test sets. Errors are mean $\pm$ sample standard deviation across paired fits. The difference is FAST minus the spectral-only reference, so negative values favor residual augmentation.}
\label{tab:app_ablation}
\small
\setlength{\tabcolsep}{4pt}
\begin{tabular}{lrrrr}
\toprule
Benchmark & Spectral only & FAST & Difference & Paired fit $95\%$ CI\\
\midrule
Navier--Stokes & $0.1108\pm0.0055$ & $0.1119\pm0.0021$ & $0.0011$ & $[-0.0033,0.0055]$\\
Darcy & $0.0876\pm0.0026$ & $0.0925\pm0.0019$ & $0.0048$ & $[0.0005,0.0092]$\\
Terminal wave & $0.2557\pm0.0016$ & $0.2063\pm0.0026$ & $-0.0494$ & $[-0.0535,-0.0453]$\\
\bottomrule
\end{tabular}
\end{minipage}
\end{center}

The fixed spectral path is sufficient for the low-rank Navier--Stokes family and is the more accurate reduced representation for Darcy. The residual path gives its clearest additional benefit for terminal wave prediction, where it does more than refine the reference. At the retained rank of six the spectral path alone reaches $0.2557$ against $0.2165$ for a plain branch, so that compression is by itself a net loss of $18.1\%$. The residual path converts it into the $0.2063$ of Table~\ref{tab:main}, a $4.7\%$ gain over the same plain branch. That gain sits in the center of the error distribution: the pooled $90$th percentile is $0.303$ for FAST against $0.291$ for Plain. Appendix~\ref{app:penalty} shows that the directional penalty is what makes the residual path useful. Appendix~\ref{app:map_diagnostics} examines the resulting map and measures how much input information the spectral path leaves behind in each case.

\section{Two controls on the source of the gain}
\label{app:controls}

The spectral-only reference of Appendix~\ref{app:ablation} separates the two paths but leaves two other readings open. The gain might come from compressing the branch at all, rather than from a basis estimated from the training inputs. It might also come from fitting fewer parameters, rather than from the representation. Two controls address these in turn. Both use the reported evaluation streams, the reported five paired model seeds, and the reported settings of Table~\ref{tab:protocol}; each changes one thing.

\paragraph{A random basis in place of the spectral one.} The first control replaces $Q_f$ by a random orthonormal matrix of the same effective rank, formed by a QR factorization of a Gaussian matrix and scaled by the same $1/\sqrt{p_f}$ convention. The rank rule, the factor scale, the block handling of auxiliary parameters, and the downstream network are untouched, and the residual path is absent, so this differs from the spectral-only reference in the basis alone.

\paragraph{A plain branch at the FAST parameter count.} The second control narrows a plain DeepONet until its trainable count is close to that of FAST-DeepONet on the same benchmark. One width sets the branch, trunk, and latent dimensions together, and it was chosen by one-epoch probes that read only the parameter count, so no accuracy entered the choice.

\begin{center}
\begin{minipage}{\linewidth}
\centering
\captionof{table}{The two controls beside the reported representations, on the same evaluation streams and paired seeds. Entries are mean $\pm$ sample standard deviation across fits. The matched plain widths and trainable counts are $20$ ($166{,}741$) for Navier--Stokes, $40$ ($174{,}481$) for Darcy, and $40$ ($174{,}401$) for terminal wave, against FAST counts of $168{,}713$, $171{,}265$, and $169{,}729$.}
\label{tab:app_controls}
\scriptsize
\setlength{\tabcolsep}{3pt}
\begin{tabular}{lccccc}
\toprule
Benchmark & Plain & Plain, matched & Random basis & Spectral only & FAST\\
\midrule
Navier--Stokes & $0.1776\pm0.0057$ & $0.2504\pm0.0224$ & $0.2731\pm0.0314$ & $0.1108\pm0.0055$ & $\mathbf{0.1119\pm0.0021}$\\
Darcy & $0.1029\pm0.0022$ & $0.1234\pm0.0038$ & $0.1524\pm0.0142$ & $0.0876\pm0.0026$ & $\mathbf{0.0925\pm0.0019}$\\
Terminal wave & $0.2165\pm0.0030$ & $0.2462\pm0.0063$ & $0.3748\pm0.0164$ & $0.2557\pm0.0016$ & $\mathbf{0.2063\pm0.0026}$\\
\bottomrule
\end{tabular}
\end{minipage}
\end{center}

Both controls are worse than a plain branch on all three benchmarks. The random basis loses to the spectral one by factors of $2.5$, $1.7$, and $1.5$, and it does not even reach the raw branch it replaces, so compressing to a low-dimensional subspace is not by itself the mechanism; the subspace has to be the one the training inputs single out. The parameter-matched plain branch is also worse than the full-width plain branch, so the same budget buys less when it is spent on a narrower network than when it is spent on the representation. The parameter reduction of Appendix~\ref{app:cost} is therefore a consequence of the construction rather than the reason it works.

\section{Sensitivity to the directional penalty}
\label{app:penalty}

The penalty weight $\lambda$ in \eqref{eq:training_objective} is the one FAST-DeepONet setting with no counterpart in Plain DeepONet, so we isolate its contribution directly. We refit FAST-DeepONet with $\lambda=0$ on all three benchmarks; everything else reproduces the fits behind Table~\ref{tab:main}. At $\lambda=0$ the penalty term is skipped, so the comparison removes the directional regularizer while leaving the representation intact. Both arms are scored identically and paired by model seed. This is a secondary ablation on an evaluation stream already opened for the main comparison. Table~\ref{tab:app_penalty} reports the outcome.

\begin{center}
\begin{minipage}{\linewidth}
\centering
\captionof{table}{Effect of the directional penalty. Errors are mean $\pm$ sample standard deviation across paired fits. The interval is for the selected $\lambda$ minus $\lambda=0$, so negative values favor the penalty. The last column counts the fits in which the selected $\lambda$ has the lower mean error.}
\label{tab:app_penalty}
\small
\setlength{\tabcolsep}{4pt}
\begin{tabular}{lcrrrc}
\toprule
Benchmark & $\lambda$ & $\lambda=0$ & Selected $\lambda$ & Paired fit $95\%$ CI & Wins\\
\midrule
Navier--Stokes & $0.01$ & $0.1127\pm0.0015$ & $0.1119\pm0.0021$ & $[-0.0032,0.0016]$ & 4/5\\
Darcy & $1$ & $0.1385\pm0.0110$ & $0.0925\pm0.0019$ & $[-0.0578,-0.0342]$ & 5/5\\
Terminal wave & $1$ & $0.2572\pm0.0079$ & $0.2063\pm0.0026$ & $[-0.0627,-0.0391]$ & 5/5\\
\bottomrule
\end{tabular}
\end{minipage}
\end{center}

The penalty reduces the mean error by $33.2\%$ on Darcy and $19.8\%$ on terminal wave. Both intervals exclude zero, the selected $\lambda$ wins at every fit, and the fieldwise advantage holds for $94$ and $100$ of the $100$ test fields respectively. The penalty also narrows the spread across fits, from $0.0110$ to $0.0019$ on Darcy and from $0.0079$ to $0.0026$ on wave. For Navier--Stokes the paired interval includes zero, and the selected value is $\lambda=0.01$, two orders of magnitude smaller than for the other two benchmarks. The tuning procedure had effectively turned the penalty off there, and the ablation agrees.

Read together with Table~\ref{tab:app_ablation}, these numbers separate the residual path from its regularizer. On terminal wave the spectral-only reference reaches $0.2557$ and the unregularized residual path reaches $0.2572$, so adding a residual path without the penalty gains nothing, and the entire improvement to $0.2063$ comes from regularizing it. On Darcy the unregularized residual path is actively harmful, raising the error from the spectral-only value of $0.0876$ to $0.1385$, and the penalty recovers most but not all of that loss. The directional penalty is therefore not a refinement of the residual path but the condition under which the residual path is useful at all.

\section{Sensitivity to the factor scale}
\label{app:scaling}

The fixed scale $c_f$ in \eqref{eq:factor_residual} sets the magnitude at which the spectral features enter the branch network. It leaves the orthogonal residual unchanged, so it acts only on the relative size of the two paths. Table~\ref{tab:protocol} reports $c_f=1$ for Darcy and $c_f=8$ for terminal wave, and this appendix measures what that choice is worth. We refit both benchmarks over $c_f\in\{1,2,4,8\}$, holding $\lambda$ at its selected value and every other setting fixed, and Table~\ref{tab:app_scaling} reports the grid. The tuning MSE is the statistic of interest, since that is the criterion by which $c_f$ was chosen on development data. The question is whether the same criterion, recomputed on the final streams, still ranks the reported value first. It is comparable across the grid because the target normalization depends on the data alone.

Navier--Stokes is excluded, for a reason worth stating precisely. Equation~\eqref{eq:block_preprocessing} defines $c_f$ as a scale on the spectral field scores alone, with $\eta_s$ appended unchanged. The released implementation applies it to the whole fixed block. The two agree exactly whenever $p_\eta=0$, which covers Darcy and terminal wave, and they agree at $c_f=1$ for any $p_\eta$. Navier--Stokes is the one reported benchmark with $p_\eta>0$, and it uses $c_f=1$, so no reported fit distinguishes the two readings. A grid over $c_f$ on that benchmark would, however, also rescale the viscosity feature and would therefore not measure the quantity \eqref{eq:block_preprocessing} defines, which is why it carries no grid.

\begin{center}
\begin{minipage}{\linewidth}
\centering
\captionof{table}{Factor-scale sensitivity, averaged over paired fits. Tuning MSE is the criterion by which $c_f$ was selected; its minimum in each benchmark is shown in bold. Test error is the mean fieldwise relative $L_2$ on the evaluation stream.}
\label{tab:app_scaling}
\setlength{\tabcolsep}{7pt}
\begin{tabular}{ccccc}
\toprule
& \multicolumn{2}{c}{Darcy} & \multicolumn{2}{c}{Terminal wave}\\
\cmidrule(lr){2-3}\cmidrule(lr){4-5}
$c_f$ & Tuning MSE & Test error & Tuning MSE & Test error\\
\midrule
1 & $\mathbf{0.0623}$ & $0.0925$ & $0.7331$ & $0.3980$\\
2 & $0.0689$ & $0.0900$ & $0.2784$ & $0.2385$\\
4 & $0.0693$ & $0.0904$ & $0.2325$ & $0.2140$\\
8 & $0.0708$ & $0.0923$ & $\mathbf{0.2168}$ & $0.2063$\\
\bottomrule
\end{tabular}
\end{minipage}
\end{center}

The two benchmarks behave differently. For terminal wave the scale is a first-order setting: the tuning MSE falls monotonically from $0.7331$ at $c_f=1$ to $0.2168$ at $c_f=8$, and the test error follows the same ordering, from $0.3980$ to $0.2063$. At $c_f=1$ the model is worse than Plain DeepONet, so this scale has to be matched to the retained subspace rather than left at a default. The selection criterion and the held-out ordering agree on the reported value.

For Darcy the scale barely matters. All four grid points lie between $0.0900$ and $0.0925$, a spread of $2.7\%$, which is smaller than the Plain-to-FAST gap on that benchmark. The tuning MSE is lowest at the reported $c_f=1$, while the lowest test error occurs at $c_f=2$ by $0.0025$. The reported value is the one the selection criterion chose, and it is not the test optimum. Taken together, $c_f$ needs care when the retained spectral subspace is small relative to the input variation, as in the rank six wave setting, and can be left at unity otherwise.

\section{Diagnostics of the effective residual map}
\label{app:map_diagnostics}

This appendix characterizes the learned residual map of the evaluation fits. No model is refitted; the retained checkpoints from Section~\ref{subsec:main_results} are reloaded and $G$ is recomputed from \eqref{eq:effective_map}. Raw row norms are excluded from the reported statistics because their magnitude is gauge dependent. They are used only to confirm that the exact normalization in \eqref{eq:row_normalization} never encounters a zero row: across every FAST evaluation fit no row of $G$ is exactly zero, and the smallest nonzero row norm is $0.265$.

\subsection{Directional concentration}

Table~\ref{tab:map_diagnostics} summarizes the column structure of the row-normalized map $\widetilde G$. For column norms $c_j=\lVert\widetilde G_{:,j}\rVert_2$, the top column statistic is the fraction of $\sum_j c_j^2$ contained in the largest $\lceil 0.01p\rceil$ columns, and the participation ratio is $\bigl(\sum_j c_j\bigr)^2/\sum_j c_j^2$. The participation ratio equals $p$ for a uniform profile and $1$ for a single active coordinate.

\begin{center}
\begin{minipage}{\linewidth}
\centering
\captionof{table}{Directional concentration of the row-normalized effective residual map (mean and range over the evaluation fits).}
\label{tab:map_diagnostics}
\setlength{\tabcolsep}{5pt}
\begin{tabular}{lrcc}
\toprule
Benchmark & $p$ & Energy in top $1\%$ of columns & Participation ratio\\
\midrule
Navier--Stokes & 8193 & $0.741\;[0.668,0.862]$ & $161.0\;[125.2,202.0]$\\
Darcy & 4096 & $0.892\;[0.857,0.931]$ & $56.8\;[50.5,60.9]$\\
Terminal wave & 4096 & $0.857\;[0.835,0.890]$ & $74.3\;[72.1,78.5]$\\
\bottomrule
\end{tabular}
\end{minipage}
\end{center}

The map is therefore far from uniform over the branch coordinates, but its participation ratio still corresponds to tens of active directions rather than a handful. The distribution of $c_j$ is heavy tailed, with a small number of columns carrying a large share of the total energy. An energy share would therefore be dominated by those few columns, so the alignment statistics below are reported as counts instead.

\subsection{Relation to the generative support}

The Darcy and wave input distributions activate a subset of ten fixed candidate disks per field, so a generative coordinate support is available for comparison, whereas Navier--Stokes has no localized scatterers and is omitted here. For each fit we take the union of the disk supports over its training fields, which covers $264$ coordinates for Darcy and $320$ for wave. We then count how many of the $100$ largest columns of $\widetilde G$ fall inside that union. Two controls accompany the comparison: the same mask transposed on the grid, and a random mask of equal size. Table~\ref{tab:map_support} reports the resulting counts.

\begin{center}
\begin{minipage}{\linewidth}
\centering
\captionof{table}{Number of the $100$ largest columns of $\widetilde G$ falling inside a mask, averaged over the evaluation fits. The chance column is the count expected if the largest columns were placed uniformly at random.}
\label{tab:map_support}
\setlength{\tabcolsep}{5pt}
\begin{tabular}{lcccc}
\toprule
Benchmark & Chance & True support & Transposed & Random\\
\midrule
Darcy & $6.4$ & $34.2\;[29,38]$ & $12.0$ & $8.6$\\
Terminal wave & $7.8$ & $18.4\;[15,23]$ & $3.6$ & $6.0$\\
\bottomrule
\end{tabular}
\end{minipage}
\end{center}

The largest columns are enriched on the true support by factors of $5.3$ and $2.4$, while both controls stay near or below chance. The residual path therefore does place above-chance weight on the coordinates where the scatterers sit. It does not, however, recover that support. Roughly two thirds of the largest columns still lie outside the disks, and the map remains spread over tens of directions. The selected coordinates also vary between fits: the Jaccard overlap of the top $100$ columns is $0.085$ for Darcy and $0.034$ for wave, against a chance value of $0.012$. Different fits of the same configuration thus use different coordinates within a partly shared region. We consequently describe the learned path as a distributed residual representation with weak support alignment.

\subsection{Residual information after the spectral path}

The same checkpoints also indicate why the residual path matters on one benchmark and not on another. For each fit we form the field residual $e_f(x_f)$ of \eqref{eq:factor_residual} and compare its per-coordinate variance with that of the input. Table~\ref{tab:map_residual_information} reports the fraction of input variance that survives the projection.

\begin{center}
\begin{minipage}{\linewidth}
\centering
\captionof{table}{Fraction of the per-coordinate training input variance retained by the orthogonal residual, averaged over the evaluation fits. The effective field rank is $16$ for Darcy and $6$ for wave. The last column is the ratio of absolute residual variance on the support to absolute residual variance elsewhere. It is not the quotient of the two preceding columns, because those are fractions of an input variance that is itself larger on the support, by $1.99$ for Darcy and $6.74$ for wave; the quotient of the fractions is $34.8$ and $6.5$ respectively.}
\label{tab:map_residual_information}
\setlength{\tabcolsep}{6pt}
\begin{tabular}{lccc}
\toprule
Benchmark & Retained on support & Retained elsewhere & Residual variance ratio\\
\midrule
Darcy & $3.8\%$ & $0.11\%$ & $69.2$\\
Terminal wave & $61.5\%$ & $9.45\%$ & $43.9$\\
\bottomrule
\end{tabular}
\end{minipage}
\end{center}

In both benchmarks the absolute residual variance sits on the scatterer coordinates, by factors of $69.2$ and $43.9$ relative to the rest of the field. What differs is how much is left there at all, and Section~\ref{subsec:mechanism} reads the two cases against the accuracy they produce.

\section{Choosing the representation parameters}
\label{app:recipe}

The settings of Table~\ref{tab:protocol} were fixed as Appendix~\ref{app:protocol} describes. This appendix collects what the sensitivity studies suggest for a new problem. The order matters: the first two steps use training inputs only and need no fit, and they determine how much the last two can matter.

\paragraph{Rank $r$, from the training inputs.} The spectral path is estimated from $X_f$ alone, so its diagnostics are available before any model is trained. Table~\ref{tab:app_spectra} reports the two that are useful: the numerical rank of \eqref{eq:rank_rule_field} and $k_{95}$, the number of centered components carrying $95\%$ of the input variance. Requesting $r$ near $k_{95}$ is a reasonable starting point, and the safeguard in \eqref{eq:rank_rule_field} removes the risk of asking for more than the data support. Navier--Stokes illustrates this: $r=8$ was requested and $6$ retained, because the training fields resolve only five field directions.

\paragraph{Residual width $s$, after checking what is left.} Whether the residual path has anything to encode is also measurable in advance. Projecting the training inputs at the chosen rank and computing the retained variance, as in Table~\ref{tab:map_residual_information}, separates the two regimes seen here. At the Darcy rank of $16$ the residual keeps $3.8\%$ of the input variance on the scatterer coordinates and the spectral-only reference is the more accurate model; at the wave rank of $6$ it keeps $61.5\%$ and the residual path lowers the error by $19.3\%$. A small retained fraction is a signal that capacity spent on $s$ will not be repaid.

\paragraph{Factor scale $c_f$, by tuning error.} The scale matters when the retained subspace is small relative to the input variation and is close to irrelevant otherwise. Table~\ref{tab:app_scaling} shows both: on wave at rank six the tuning MSE falls from $0.7331$ at $c_f=1$ to $0.2168$ at $c_f=8$ and the held-out ordering agrees, while on Darcy at rank sixteen the four grid points span $2.7\%$. Selecting by mean tuning MSE reproduced the reported value on both. Leaving $c_f=1$ is safe when the retained rank is generous, and is the only consistent choice when the branch carries auxiliary parameters.

\paragraph{Penalty weight $\lambda$, never zero.} Appendix~\ref{app:penalty} is unambiguous on this point: an unregularized residual path gains nothing on wave and is actively harmful on Darcy. Selecting $\lambda$ by mean tuning MSE is what produced the reported values, and it also produced the sensible outcome of a near-inactive penalty on Navier--Stokes, where little residual variance remains to regularize. The clipping threshold $\tau_k$ of \eqref{eq:tau_schedule} was fixed a priori, so it was never tuned against an evaluation stream.

These steps summarize the behavior observed across the three benchmarks.

\section{Fixed optimization-budget diagnostic}
\label{app:fixed_budget}

To assess sensitivity to a shorter optimization budget, we repeated one prespecified Plain/FAST seed per benchmark under a common cap of $200$ epochs, with all other settings matching the final protocol. The penalty schedule is defined over this run, so it is a separate experiment and not a prefix of the longer fits. Table~\ref{tab:app_fixed_budget} reports all three outcomes.

\begin{center}
\begin{minipage}{\linewidth}
\centering
\captionof{table}{Fixed-budget diagnostic using the first paired seed of each benchmark. Each entry is the mean fieldwise relative $L_2$ error; the retained epoch minimizes tuning MSE within the common budget.}
\label{tab:app_fixed_budget}
\setlength{\tabcolsep}{5pt}
\begin{tabular}{lrrrrr}
\toprule
Benchmark & Plain & FAST & Reduction & Plain epoch & FAST epoch\\
\midrule
Navier--Stokes & $0.1737$ & $0.1176$ & $32.3\%$ & 71 & 200\\
Darcy & $0.1063$ & $0.0947$ & $10.9\%$ & 189 & 200\\
Terminal wave & $0.2153$ & $0.2140$ & $0.6\%$ & 139 & 198\\
\bottomrule
\end{tabular}
\end{minipage}
\end{center}

Both Navier--Stokes and Darcy retain a clear reduction under the shorter budget. For terminal wave the two models are nearly tied, and the FAST checkpoint is still improving at the cap, so the wave result in Table~\ref{tab:main} depends on the longer schedule.

\section{Training curves}
\label{app:training_curves}

No model is refitted here. Figure~\ref{fig:training_curves} draws the per-epoch tuning MSE that the runners stored for the fits behind Table~\ref{tab:main}, so the curves describe the models that produced the reported errors. Each panel carries the five paired model seeds of both representations.

\begin{center}
\begin{minipage}{\linewidth}
\centering
\includegraphics[width=\linewidth]{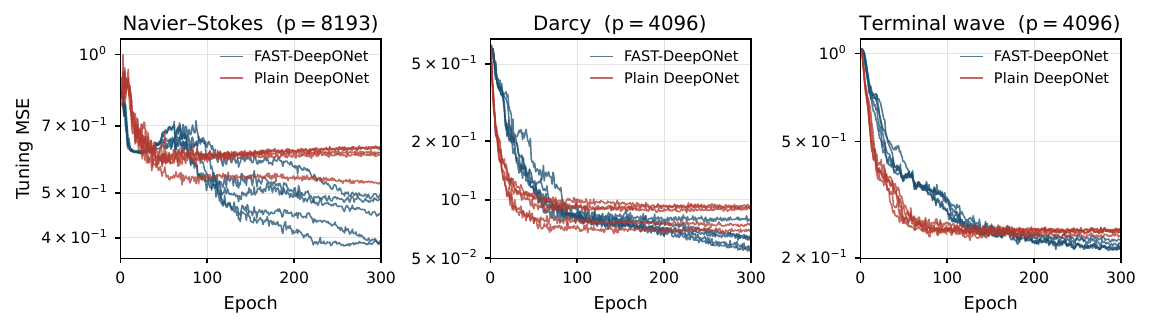}
\captionof{figure}{Tuning MSE against the epoch for the five paired model seeds of each representation, where lower is better. The vertical range is set separately in each panel, so the levels are comparable within a benchmark and not across benchmarks.}
\label{fig:training_curves}
\end{minipage}
\end{center}

Two readings matter. First, neither representation is stopped before its own optimum: the checkpoint rule retains the epoch of lowest tuning MSE, and that epoch lies inside the budget for every fit. The reported plain errors are therefore the best that branch reaches under this schedule.

Second, the two branches leave the descent at different times. On Navier--Stokes the retained epoch is $37$ to $71$ for Plain and $222$ to $297$ for FAST, and the plain tuning MSE is $6.3\%$ above its minimum by the final epoch against $1.1\%$ for FAST. The plain branch stops improving early and then drifts back, which is the behavior of a first layer that the training operators do not determine. The same ordering holds on the two smaller benchmarks with a smaller gap.

\section{Parameter count and training cost}
\label{app:cost}

Replacing the raw branch by the factor and residual representation changes the model size sharply, because the first branch layer no longer reads every sensor. Table~\ref{tab:app_cost} reports counts taken from the retained checkpoints, so they describe the models that produced the reported errors.

\begin{center}
\begin{minipage}{\linewidth}
\centering
\captionof{table}{Model size. The branch input dimension is $r_{\mathrm{eff}}+s$ for FAST and $p$ for Plain. Counts are taken from the retained checkpoints; the fixed buffer stores $Q_f$ and its reconstruction and carries no gradient.}
\label{tab:app_cost}
\setlength{\tabcolsep}{6pt}
\begin{tabular}{llrrr}
\toprule
Benchmark & Model & Branch input & Trainable & Fixed buffer\\
\midrule
\multirow{2}{*}{Navier--Stokes}
 & Plain & 8193 & $1{,}150{,}081$ & $0$\\
 & FAST & 14 & $\mathbf{168{,}713}$ & $98{,}316$\\
\addlinespace[2pt]
\multirow{2}{*}{Darcy}
 & Plain & 4096 & $625{,}921$ & $0$\\
 & FAST & 32 & $\mathbf{171{,}265}$ & $131{,}072$\\
\addlinespace[2pt]
\multirow{2}{*}{Terminal wave}
 & Plain & 4096 & $625{,}665$ & $0$\\
 & FAST & 22 & $\mathbf{169{,}729}$ & $49{,}152$\\
\bottomrule
\end{tabular}
\end{minipage}
\end{center}

FAST-DeepONet trains $6.8$ times fewer parameters than Plain on Navier--Stokes and $3.7$ times fewer on each of the two smaller benchmarks. The reduction is structural: the Plain branch spends most of its parameters on a first layer of size $w\times p$, which the representation replaces by $w\times(r_{\mathrm{eff}}+s)$.

Training time moves the other way, but by much less. FAST-DeepONet forms the $w\times p$ effective map at every optimizer step, which brings one fit to about $1.3$ times the wall-clock of a plain branch. That cost grows linearly in $p$ and does not depend on the number of training fields.

\section{Sensor resolution and sample size}
\label{app:pn_full}

This appendix gives the four Navier--Stokes corners behind the reproducibility statement of Section~\ref{subsec:pn}, the construction of the two fresh training banks, every fitted cell of Figure~\ref{fig:pn_sweep_all}, and the reason the designed benchmarks start at $16\times16$.

Figure~\ref{fig:pn} shows the four Navier--Stokes corners, each with five paired model seeds. Interior cells of that grid use one prespecified seed and are descriptive. All Navier--Stokes training subsets are nested within one fixed $240$ field bank.

\begin{center}
\begin{minipage}{\linewidth}
\centering
\includegraphics[width=0.72\linewidth]{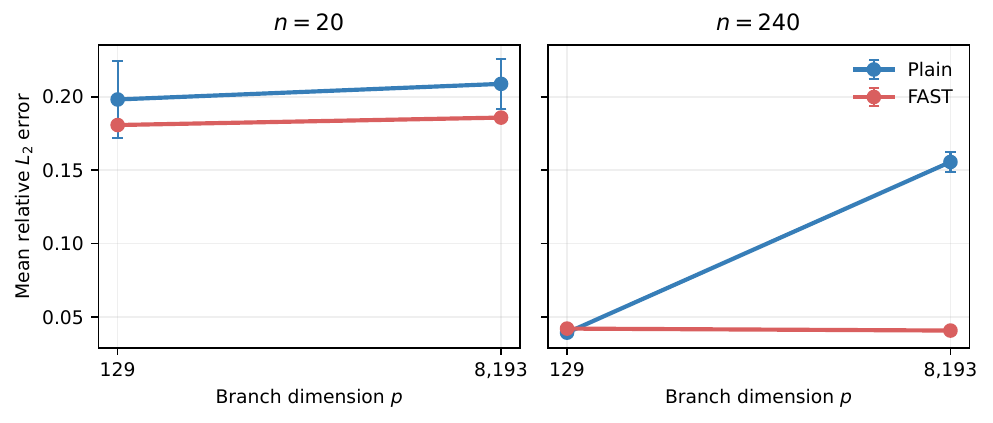}
\captionof{figure}{Navier--Stokes results at the four corners of the sensor resolution and sample size study, where lower is better. Markers and lines show means across paired fits; whiskers show one sample standard deviation. The target grid remains $128\times128$ in every case.}
\label{fig:pn}
\end{minipage}
\end{center}

\subsection{The designed benchmarks}

Darcy and terminal wave need a training bank larger than the reported split before the same sweep can be run. For Darcy the generator stream is continued from index $610$, since indices $0$ to $609$ are already allocated to the reported study; the regenerated stream reproduces files on disk at seven probe indices before anything new is written. For terminal wave the confirmatory generator supplies fresh per-sample seeds, because the earlier stream predates the zero-velocity half-step initialization used by the reported benchmark. Both banks carry the candidate sites of their benchmark and share no field with any reported split, and both sweeps are scored on the reported evaluation stream. Training subsets are nested, and the queries per epoch scale with the number of training fields.

The sweep draws its subsets from banks disjoint from the reported training splits, so its cells are not numerically comparable with Table~\ref{tab:main}. They are comparable with each other.

Table~\ref{tab:app_pn_sweep} gives every fitted cell. That the wave column stays below the other two matches the reading of Appendix~\ref{app:ablation}: at the retained rank of six the spectral path alone is a net loss against a plain branch, and only the complete representation recovers it.

\begin{center}
\begin{minipage}{\linewidth}
\centering
\captionof{table}{FAST reduction in mean relative $L_2$ error against a plain branch at the same sensor grid and training set size. Positive values favor FAST. Reading down a column gives the dependence on the branch dimension at a fixed sample size.}
\label{tab:app_pn_sweep}
\small
\setlength{\tabcolsep}{4pt}
\begin{tabular}{lrrrrrrrrr}
\toprule
& \multicolumn{3}{c}{Navier--Stokes} & \multicolumn{3}{c}{Darcy} & \multicolumn{3}{c}{Terminal wave}\\
\cmidrule(lr){2-4}\cmidrule(lr){5-7}\cmidrule(lr){8-10}
Sensor grid & $n=20$ & $60$ & $240$ & $n=20$ & $60$ & $240$ & $n=20$ & $60$ & $240$\\
\midrule
$8\times8$ & $+8.8$ & $+14.7$ & $-7.1$ & $+5.5$ & $-2.6$ & $+12.3$ & $+1.0$ & $-22.0$ & $-1.1$\\
$16\times16$ & $+17.7$ & $+42.4$ & $+8.6$ & $+1.8$ & $+0.9$ & $+16.0$ & $-3.2$ & $-9.3$ & $-34.3$\\
$32\times32$ & $+4.8$ & $+53.1$ & $+54.6$ & $+1.9$ & $+4.0$ & $+21.4$ & $-5.0$ & $-4.1$ & $-16.6$\\
$64\times64$ & $+11.0$ & $+56.0$ & $+73.8$ & $+9.2$ & $+10.5$ & $+31.1$ & $-1.0$ & $-1.2$ & $+5.6$\\
\bottomrule
\end{tabular}
\end{minipage}
\end{center}

\subsection{What a coarse sensor grid measures}

Coarsening the grid is point sampling, not averaging, so a scatterer narrower than the sensor spacing can fall between sensors. The disks have diameter $0.09$ for Darcy and $0.10$ for wave, against a spacing of $0.125$ at $8\times8$ and $0.0625$ at $16\times16$. Counting on the training banks, an $8\times8$ grid places a sensor inside only $30.9\%$ of the active Darcy disks and $8.5\%$ of the wave disks, while from $16\times16$ onwards every active disk contains at least one sensor. The $8\times8$ row therefore samples an input distribution with most of the localized structure absent rather than a coarser view of the same one, which is why Figure~\ref{fig:pn_sweep_all} starts at $16\times16$ while Table~\ref{tab:app_pn_sweep} keeps the row.

\subsection{The reported wave configuration on a fresh bank}

The sweep grid does not contain the $40$ training fields of the reported wave study, so that cell was fitted separately on the fresh bank at the reported sensor grid, sample size, and query budget. Across five paired seeds FAST-DeepONet reaches $0.2206\pm0.0029$ against $0.2357\pm0.0037$ for a plain branch, a $6.4\%$ reduction with paired interval $[-0.0211,-0.0091]$ and the lower error at every seed. Table~\ref{tab:main} reports $4.7\%$ on the confirmatory bank, so the wave result reproduces on training fields that entered no part of the reported study.

\section{Terminal wave propagation}
\label{app:wave_propagation}

Figure~\ref{fig:app_wave_propagation} follows the same heterogeneous medium shown in Figure~\ref{fig:benchmarks}(c). The squared slowness panel is not repeated here because it already appears in the benchmark figure.

\begin{center}
\begin{minipage}{\linewidth}
\centering
\includegraphics[width=\linewidth]{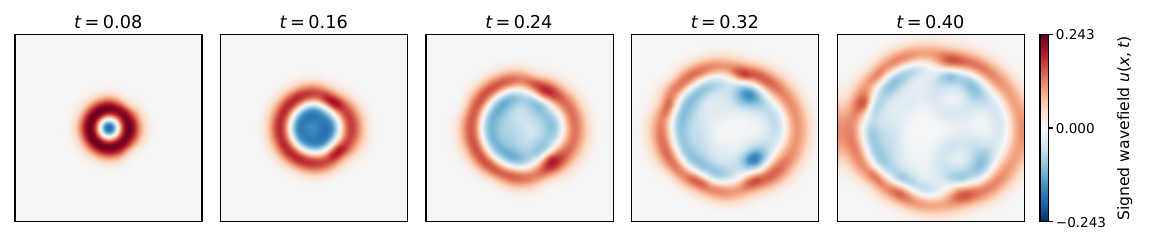}
\captionof{figure}{Signed wavefield snapshots from $t=0.08$ to the terminal time $T=0.4$. The centered pulse propagates through the heterogeneous medium and develops distorted wave fronts. All panels use one symmetric color scale, and the final panel is the learning target.}
\label{fig:app_wave_propagation}
\end{minipage}
\end{center}

\section{Monte Carlo dropout and operator-level conformal calibration}
\label{app:conformal}

We evaluate an optional uncertainty variant in which both Plain and FAST are retrained with dropout rate $0.1$ after every ReLU hidden layer of the branch and trunk \citep{gal2016dropout}. The data, seeds, schedule, and checkpoint rule match the main study. At inference, the point prediction is the mean of ten Monte Carlo dropout passes with a paired random-number schedule. These dropout fits are separate from the deterministic point predictors in Tables~\ref{tab:main} and \ref{tab:contrasts}.

We then apply split conformal calibration \citep{vovk2005algorithmic,lei2018distribution} as a secondary reliability analysis \citep{angelopoulos2023conformal,moya2025conformalized,ma2024calibrated}. The dropout architecture, fitting protocol, and all calibration choices were fixed before generating the fresh calibration and UQ test fields listed in Table~\ref{tab:app_settings}. The absolute residual at query $j$ of calibration operator $i$ is
\begin{equation}
    R_{ij}=|y_{ij}-\overline y_{ij}^{\mathrm{MC}}|,
    \label{eq:app_score}
\end{equation}
where $\overline y^{\mathrm{MC}}$ is the ten-pass mean. With $m_i$ native grid points and $\beta=0.1$, define the within-operator score
\begin{equation}
    k_{\beta,i}=\min\{m_i,\lceil(m_i+1)(1-\beta)\rceil\},
    \qquad Q_i=R_{i,(k_{\beta,i})}.
    \label{eq:app_inner}
\end{equation}
For $K=20$ calibration operators and $\alpha=0.1$, let
\begin{equation}
    k_\alpha=\min\{K,\lceil(K+1)(1-\alpha)\rceil\},
    \qquad \widehat q=Q_{(k_\alpha)}.
    \label{eq:app_outer}
\end{equation}
The interval at a new query is $[\overline y^{\mathrm{MC}}-\widehat q,\overline y^{\mathrm{MC}}+\widehat q]$. If all choices are fixed before calibration and the calibration operators and a new operator are exchangeable, the split conformal rank argument gives
\begin{equation}
    \Pr\!\left[
      \frac{1}{m_*}\sum_{j=1}^{m_*}
      \mathbf 1\{R_{*,j}\le\widehat q\}
      \ge \frac{k_{\beta,*}}{m_*}
    \right]\ge 1-\alpha.
    \label{eq:app_guarantee}
\end{equation}
This is an operator-level statement about the fraction of covered grid points, marginal over the random calibration sample and a new operator.

Table~\ref{tab:app_conformal} reports the independent empirical study. FAST has lower MC-mean point error on all three benchmarks. Its intervals are shorter on Navier--Stokes and Darcy, while the terminal-wave intervals are slightly wider. Field success behaves differently on each benchmark. All five FAST seeds are flagged on Navier--Stokes, both methods are flagged on Darcy, and neither method is flagged on terminal wave. A flag is a descriptive statement about one realized calibration set. The guarantee in \eqref{eq:app_guarantee} is marginal over the calibration draw and a new operator, so a flag does not contradict it and is not a test of it.

\begin{center}
\begin{minipage}{\linewidth}
\centering
\captionof{table}{Operator-level conformal evaluation of the dropout variants. MC error is measured on the aligned accuracy stream; width and coverage metrics use the independent fresh UQ stream. Entries are mean $\pm$ sample standard deviation across fitted seeds. Field success is the fraction of UQ fields attaining the spatial target in \eqref{eq:app_inner}; brackets give the range across seeds. A flag is a conditional empirical diagnostic: the upper endpoint of that seed's $95\%$ Wilson interval, computed on one realized calibration set, lies below the nominal $0.9$ field-success target. Width is in original target units.}
\label{tab:app_conformal}
\scriptsize
\setlength{\tabcolsep}{2.4pt}
\begin{tabular}{llcccccc}
\toprule
Benchmark & Model & MC error & Width & Mean cov. & P10 cov. & Success range & Flags\\
\midrule
\multirow{2}{*}{Navier--Stokes}
 & Plain & $0.1675\pm0.0029$ & $0.5115\pm0.0152$ & $0.9646\pm0.0050$ & $0.8843\pm0.0226$ & $0.85$--$0.90$ & 0/5\\
 & FAST & $\mathbf{0.1258\pm0.0036}$ & $\mathbf{0.3537\pm0.0199}$ & $0.9438\pm0.0071$ & $0.8091\pm0.0325$ & $0.78$--$0.82$ & 5/5\\
\addlinespace[2pt]
\multirow{2}{*}{Darcy}
 & Plain & $0.1138\pm0.0024$ & $0.01868\pm0.00138$ & $0.9333\pm0.0087$ & $0.8367\pm0.0182$ & $0.68$--$0.77$ & 5/5\\
 & FAST & $\mathbf{0.1071\pm0.0025}$ & $\mathbf{0.01632\pm0.00087}$ & $0.9240\pm0.0080$ & $0.8208\pm0.0172$ & $0.63$--$0.73$ & 5/5\\
\addlinespace[2pt]
\multirow{2}{*}{Terminal wave}
 & Plain & $0.2331\pm0.0029$ & $\mathbf{0.05853\pm0.00224}$ & $0.9442\pm0.0049$ & $0.9089\pm0.0069$ & $0.89$--$0.96$ & 0/5\\
 & FAST & $\mathbf{0.2250\pm0.0016}$ & $0.06079\pm0.00288$ & $0.9560\pm0.0055$ & $0.9226\pm0.0076$ & $0.93$--$0.98$ & 0/5\\
\bottomrule
\end{tabular}
\end{minipage}
\end{center}

\section{Implementation checks}
\label{app:checks}

The exact row normalization, residual coordinate gauge, factor null-space gauge, and positive ReLU row scaling were tested directly. The effective map and penalty remain unchanged to floating-point tolerance under the corresponding transformations, while zero-row backward evaluations remain finite.

We also performed a solver-only audit at five fixed accuracy-test indices spanning each stream, without loading or evaluating a learned model. Table~\ref{tab:app_solver} reports float64 solver replay and refinement results. The refinement error is the relative $L_2$ difference between the production solution and the refined solution restricted to the production grid.

\begin{center}
\begin{minipage}{\linewidth}
\centering
\captionof{table}{Solver verification on five fixed inputs. Replay and solver-check entries are means unless marked as maxima; refinement entries are mean (maximum). Navier--Stokes jointly refines $128^2$, $\Delta t=0.001$ to $256^2$, $\Delta t=0.0005$; Darcy refines $64^2$ to $128^2$; wave refines $64^2$, $\Delta t=0.004$ to $128^2$, $\Delta t=0.002$.}
\label{tab:app_solver}
\small
\setlength{\tabcolsep}{4pt}
\begin{tabular}{lccc}
\toprule
Benchmark & Target replay & Solver check & Refinement relative $L_2$\\
\midrule
Navier--Stokes & $2.43\times10^{-8}$ & div. RMS $2.90\times10^{-14}$ & $6.23\times10^{-4}$ ($7.24\times10^{-4}$)\\
Darcy & $2.54\times10^{-8}$ & linear residual $1.61\times10^{-13}$ & $1.60\times10^{-2}$ ($2.91\times10^{-2}$)\\
Terminal wave & $2.76\times10^{-8}$ & max CFL $0.323<0.707$ & $4.25\times10^{-2}$ ($4.84\times10^{-2}$)\\
\bottomrule
\end{tabular}
\end{minipage}
\end{center}

The Navier--Stokes divergence and Darcy linear residual are evaluated on the float64 replay before serialization; target replay quantifies the subsequent float32 storage difference. Darcy refinement also changes the grid representation of discontinuous disk interfaces. For wave, regeneration from the saved seed reproduces the stored arrays exactly. The table instead reports re-solving the stored medium, with the refined periodic field sampled at the coarse cell centers. These checks support numerical consistency of the generators.

\bibliographystyle{unsrtnat}
\bibliography{references}

@article{fan2024fastnn,
  author = {Fan, Jianqing and Gu, Yihong},
  title = {Factor Augmented Sparse Throughput Deep ReLU Neural Networks for High Dimensional Regression},
  journal = {Journal of the American Statistical Association},
  volume = {119},
  number = {548},
  pages = {2680--2694},
  year = {2024},
  doi = {10.1080/01621459.2023.2271605},
  url = {https://doi.org/10.1080/01621459.2023.2271605}
}

@article{lu2021deeponet,
  author = {Lu, Lu and Jin, Pengzhan and Pang, Guofei and Zhang, Zhongqiang and Karniadakis, George Em},
  title = {Learning nonlinear operators via DeepONet based on the universal approximation theorem of operators},
  journal = {Nature Machine Intelligence},
  volume = {3},
  number = {3},
  pages = {218--229},
  year = {2021},
  doi = {10.1038/s42256-021-00302-5},
  url = {https://doi.org/10.1038/s42256-021-00302-5}
}

@article{lu2022comprehensive,
  author = {Lu, Lu and Meng, Xuhui and Cai, Shengze and Mao, Zhiping and Goswami, Somdatta and Zhang, Zhongqiang and Karniadakis, George Em},
  title = {A Comprehensive and Fair Comparison of Two Neural Operators (with Practical Extensions) Based on FAIR Data},
  journal = {Computer Methods in Applied Mechanics and Engineering},
  volume = {393},
  pages = {114778},
  year = {2022},
  doi = {10.1016/j.cma.2022.114778},
  eprint = {2111.05512},
  archivePrefix = {arXiv},
  primaryClass = {cs.LG},
  url = {https://doi.org/10.1016/j.cma.2022.114778}
}

@inproceedings{gal2016dropout,
  author = {Gal, Yarin and Ghahramani, Zoubin},
  title = {Dropout as a Bayesian Approximation: Representing Model Uncertainty in Deep Learning},
  booktitle = {Proceedings of the 33rd International Conference on Machine Learning},
  series = {Proceedings of Machine Learning Research},
  volume = {48},
  pages = {1050--1059},
  year = {2016},
  url = {https://proceedings.mlr.press/v48/gal16.html}
}

@book{vovk2005algorithmic,
  author = {Vovk, Vladimir and Gammerman, Alexander and Shafer, Glenn},
  title = {Algorithmic Learning in a Random World},
  publisher = {Springer},
  address = {New York},
  year = {2005},
  doi = {10.1007/b106715},
  url = {https://doi.org/10.1007/b106715}
}

@article{lei2018distribution,
  author = {Lei, Jing and G'Sell, Max and Rinaldo, Alessandro and Tibshirani, Ryan J. and Wasserman, Larry},
  title = {Distribution-Free Predictive Inference for Regression},
  journal = {Journal of the American Statistical Association},
  volume = {113},
  number = {523},
  pages = {1094--1111},
  year = {2018},
  doi = {10.1080/01621459.2017.1307116},
  url = {https://doi.org/10.1080/01621459.2017.1307116}
}

@inproceedings{li2021fourier,
  author = {Li, Zongyi and Kovachki, Nikola and Azizzadenesheli, Kamyar and Liu, Burigede and Bhattacharya, Kaushik and Stuart, Andrew and Anandkumar, Anima},
  title = {Fourier Neural Operator for Parametric Partial Differential Equations},
  booktitle = {International Conference on Learning Representations},
  year = {2021},
  url = {https://openreview.net/forum?id=c8P9NQVtmnO}
}

@article{kovachki2023neural,
  author = {Kovachki, Nikola and Li, Zongyi and Liu, Burigede and Azizzadenesheli, Kamyar and Bhattacharya, Kaushik and Stuart, Andrew and Anandkumar, Anima},
  title = {Neural Operator: Learning Maps Between Function Spaces With Applications to {PDE}s},
  journal = {Journal of Machine Learning Research},
  volume = {24},
  number = {89},
  pages = {1--97},
  year = {2023}
}

@article{bhattacharya2021model,
  author = {Bhattacharya, Kaushik and Hosseini, Bamdad and Kovachki, Nikola B. and Stuart, Andrew M.},
  title = {Model Reduction And Neural Networks For Parametric {PDE}s},
  journal = {The SMAI Journal of Computational Mathematics},
  volume = {7},
  pages = {121--157},
  year = {2021},
  doi = {10.5802/smai-jcm.74}
}

@article{moya2025conformalized,
  author = {Moya, Christian and Mollaali, Amirhossein and Zhang, Zecheng and Lu, Lu and Lin, Guang},
  title = {Conformalized-{DeepONet}: A distribution-free framework for uncertainty quantification in deep operator networks},
  journal = {Physica D: Nonlinear Phenomena},
  volume = {471},
  pages = {134418},
  year = {2025},
  doi = {10.1016/j.physd.2024.134418}
}

@article{angelopoulos2023conformal,
  author = {Angelopoulos, Anastasios N. and Bates, Stephen},
  title = {Conformal Prediction: A Gentle Introduction},
  journal = {Foundations and Trends in Machine Learning},
  volume = {16},
  number = {4},
  pages = {494--591},
  year = {2023},
  doi = {10.1561/2200000101}
}

@article{fan2022learning,
  author = {Fan, Jianqing and Liao, Yuan},
  title = {Learning Latent Factors From Diversified Projections and Its Applications to Over-Estimated and Weak Factors},
  journal = {Journal of the American Statistical Association},
  volume = {117},
  number = {538},
  pages = {909--924},
  year = {2022},
  doi = {10.1080/01621459.2020.1831927}
}

@article{kontolati2024learning,
  author  = {Kontolati, Katiana and Goswami, Somdatta and Karniadakis, George Em and Shields, Michael D.},
  title   = {Learning nonlinear operators in latent spaces for real-time predictions of complex dynamics in physical systems},
  journal = {Nature Communications},
  volume  = {15},
  pages   = {5101},
  year    = {2024},
  doi     = {10.1038/s41467-024-49411-w}
}

@article{bai2002determining,
  author  = {Bai, Jushan and Ng, Serena},
  title   = {Determining the number of factors in approximate factor models},
  journal = {Econometrica},
  volume  = {70},
  number  = {1},
  pages   = {191--221},
  year    = {2002},
  doi     = {10.1111/1468-0262.00273}
}

@article{fan2020farm,
  author  = {Fan, Jianqing and Ke, Yuan and Wang, Kaizheng},
  title   = {Factor-adjusted regularized model selection},
  journal = {Journal of Econometrics},
  volume  = {216},
  number  = {1},
  pages   = {71--85},
  year    = {2020},
  doi     = {10.1016/j.jeconom.2020.01.006}
}

@article{chen1995universal,
  author  = {Chen, Tianping and Chen, Hong},
  title   = {Universal approximation to nonlinear operators by neural networks with arbitrary activation functions and its application to dynamical systems},
  journal = {IEEE Transactions on Neural Networks},
  volume  = {6},
  number  = {4},
  pages   = {911--917},
  year    = {1995},
  doi     = {10.1109/72.392253}
}

@article{lanthaler2022error,
  author  = {Lanthaler, Samuel and Mishra, Siddhartha and Karniadakis, George Em},
  title   = {Error estimates for {DeepONets}: a deep learning framework in infinite dimensions},
  journal = {Transactions of Mathematics and Its Applications},
  volume  = {6},
  number  = {1},
  pages   = {tnac001},
  year    = {2022},
  doi     = {10.1093/imatrm/tnac001}
}

@article{stock2002forecasting,
  author  = {Stock, James H. and Watson, Mark W.},
  title   = {Forecasting using principal components from a large number of predictors},
  journal = {Journal of the American Statistical Association},
  volume  = {97},
  number  = {460},
  pages   = {1167--1179},
  year    = {2002},
  doi     = {10.1198/016214502388618960}
}

@article{fan2013poet,
  author  = {Fan, Jianqing and Liao, Yuan and Mincheva, Martina},
  title   = {Large covariance estimation by thresholding principal orthogonal complements},
  journal = {Journal of the Royal Statistical Society: Series B},
  volume  = {75},
  number  = {4},
  pages   = {603--680},
  year    = {2013},
  doi     = {10.1111/rssb.12016}
}

@inproceedings{seidman2022nomad,
  author    = {Seidman, Jacob H. and Kissas, Georgios and Perdikaris, Paris and Pappas, George J.},
  title     = {{NOMAD}: Nonlinear manifold decoders for operator learning},
  booktitle = {Advances in Neural Information Processing Systems},
  volume    = {35},
  year      = {2022}
}

@article{hesthaven2018nonintrusive,
  author  = {Hesthaven, Jan S. and Ubbiali, Stefano},
  title   = {Non-intrusive reduced order modeling of nonlinear problems using neural networks},
  journal = {Journal of Computational Physics},
  volume  = {363},
  pages   = {55--78},
  year    = {2018},
  doi     = {10.1016/j.jcp.2018.02.037}
}

@article{fan2001variable,
  author  = {Fan, Jianqing and Li, Runze},
  title   = {Variable selection via nonconcave penalized likelihood and its oracle properties},
  journal = {Journal of the American Statistical Association},
  volume  = {96},
  number  = {456},
  pages   = {1348--1360},
  year    = {2001},
  doi     = {10.1198/016214501753382273}
}

@article{ma2024calibrated,
  title={Calibrated uncertainty quantification for operator learning via conformal prediction},
  author={Ma, Ziqi and Pitt, David and Azizzadenesheli, Kamyar and Anandkumar, Anima},
  journal={Transactions on Machine Learning Research},
  year={2024},
  url={https://openreview.net/forum?id=cGpegxy12T}
}

@inproceedings{qiu2024derivative,
  author    = {Qiu, Yuan and Bridges, Nolan and Chen, Peng},
  title     = {Derivative-enhanced Deep Operator Network},
  booktitle = {Advances in Neural Information Processing Systems},
  volume    = {37},
  pages     = {20945--20981},
  year      = {2024},
  doi       = {10.52202/079017-0660},
  url       = {https://proceedings.neurips.cc/paper_files/paper/2024/hash/25297252dcd3f4b11eec7ec7ab06bc80-Abstract-Conference.html}
}

@inproceedings{neyshabur2015pathsgd,
  author    = {Neyshabur, Behnam and Salakhutdinov, Ruslan and Srebro, Nati},
  title     = {{Path-SGD}: Path-Normalized Optimization in Deep Neural Networks},
  booktitle = {Advances in Neural Information Processing Systems},
  volume    = {28},
  year      = {2015},
  url       = {https://proceedings.neurips.cc/paper/2015/hash/eaa32c96f620053cf442ad32258076b9-Abstract.html}
}

\end{document}